\documentclass[a4paper,fleqn]{cas-dc}

\usepackage[authoryear,longnamesfirst]{natbib}
\usepackage{lineno}
\modulolinenumbers[5]
\usepackage{graphicx} 
\usepackage[font=small,labelfont=bf]{caption}
\usepackage{subcaption}

\usepackage{algorithm}
\usepackage{algpseudocode}
\usepackage{hyperref} 

\def\tsc#1{\csdef{#1}{\textsc{\lowercase{#1}}\xspace}}
\tsc{WGM}
\tsc{QE}
\tsc{EP}
\tsc{PMS}
\tsc{BEC}
\tsc{DE}

\begin{document}
\let\WriteBookmarks\relax
\def\floatpagepagefraction{1}
\def\textpagefraction{.001}

\shorttitle{Blockchain anomalous transactions detection}

\shortauthors{M Hasan et~al.}

\title [mode = title]{Detecting Anomalies in Blockchain Transactions using Machine Learning Classifiers and Explainability Analysis}  




%
\author[1]{Mohammad Hasan}






\affiliation[1]{organization={Department of Computer Science and Engineering, Premier University},
    city={Chitagong},
    postcode={4000}, 
    country={Bangladesh}}

\author[2]{Mohammad Shahriar Rahman}


\affiliation[2]{organization={Department of Computer Science and Engineering, United International University},
    city={Dhaka},
    country={Bangladesh}}

\author[3,4]{Helge Janicke}

\author[3,4]{Iqbal H. Sarker}

\affiliation[3]{organization={Cyber Security Cooperative Research Centre},
    city={Perth},
    postcode={6027}, 
    state={WA},
    country={Australia}}

\affiliation[4]{organization={Security Research Institute, Edith Cowan University},
    city={Perth},
    postcode={6027}, 
    state={WA},
    country={Australia}}

\fnmark[*]
\ead{m.sarker@ecu.edu.au}
\ead[ORCID]{https://orcid.org/0000-0003-1740-5517}
\cortext[cor1]{Corresponding author}



\begin{abstract}
As the use of Blockchain for digital payments continues to rise in popularity, it also becomes susceptible to various malicious attacks. Successfully detecting anomalies within Blockchain transactions is essential for bolstering trust in digital payments. However, the task of anomaly detection in Blockchain transaction data is challenging due to the infrequent occurrence of illicit transactions. Although several studies have been conducted in the field, a limitation persists: the lack of explanations for the model's predictions. This study seeks to overcome this limitation by integrating eXplainable Artificial Intelligence (XAI) techniques and anomaly rules into tree-based ensemble classifiers for detecting anomalous Bitcoin transactions. The Shapley Additive exPlanation (SHAP) method is employed to measure the contribution of each feature, and it is compatible with ensemble models. Moreover, we present rules for interpreting whether a Bitcoin transaction is anomalous or not. Additionally, we have introduced an under-sampling algorithm named XGBCLUS, designed to balance anomalous and non-anomalous transaction data. This algorithm is compared against other commonly used under-sampling and over-sampling techniques. Finally, the outcomes of various tree-based single classifiers are compared with those of stacking and voting ensemble classifiers. Our experimental results demonstrate that: (i) XGBCLUS enhances TPR and ROC-AUC scores compared to state-of-the-art under-sampling and over-sampling techniques, and (ii) our proposed ensemble classifiers outperform traditional single tree-based machine learning classifiers in terms of accuracy, TPR, and FPR scores.
\end{abstract}




\maketitle

\section{Introduction}

Blockchain, a chain of blocks that contain the history of several transactions or records of other applications in a public ledger, has been considered an emerging technology both in academic and industrial areas since the last decade \cite{ir1}. Bitcoin, the first digital cryptocurrency, was proposed in 2008 and then successfully implemented by Satoshi Nakamoto \cite{ir3}. Although Blockchain was created to support the popular bitcoin currency, transactions of other digital crypto-currencies such as Ethereum, Ripple, Litecoin, etc., health records, transportation, IoT applications, etc. \cite{ir2} are stored in the blocks in a decentralized manner and are managed without the help of a third party organization. Prominent attributes like trustworthiness, verifiability, decentralization, and immutability have rendered Blockchain an effective integration partner for various Information and Communication Technology (ICT) applications. Nevertheless, this technology remains susceptible to an array of challenges, encompassing security breaches, privacy concerns, energy consumption, regulatory policies, and issues like selfish mining \cite{ir4}.

Despite its growing popularity in digital payments, Bitcoin remains susceptible to a range of attacks, encompassing temporal attacks, spatial attacks, and logical-partitioning attacks \cite{ir5}. To ensure the effective implementation of blockchain technology, the timely detection of malicious behavior or transactions within the network, or the identification of novel instances in the data, is imperative \cite{ir6}. Swift and appropriate actions must be taken to mitigate potential risks. Within the context of blockchain systems, anomaly detection assumes paramount importance. This facet aids in the identification and prevention of potential malicious activities, thereby upholding the system's integrity \cite{ir7}. Nonetheless, the inherent imbalance between normal and anomalous data in blockchain datasets, such as those of Bitcoin or Ethereum transactions, presents a substantial challenge for conventional anomaly detection methodologies. In many instances, the frequency of anomalous data points significantly pales in comparison to that of normal data points, thereby yielding imbalanced datasets. Such skewed data distribution can exert an adverse impact on the efficacy of anomaly detection algorithms. These algorithms, often biased towards the majority class (normal data), grapple with difficulties in accurately discerning the minority class (anomalous data) \cite{ir8}.

Several Over and Under-sampling techniques such as Synthetic Minority OverSampling Technique (SMOTE), Adaptive Synthetic (ADASYN), Random Under Sampling (RUS), Near-Miss, etc. have been used to handle imbalanced data in various domains\cite{ir16}. Under-sampling techniques can alleviate the bias towards the majority class and enhance the performance of anomaly detection algorithms \cite{ir15}. In this study, the anomalous Bitcoin transactions are much lower than the legal transactions. However, it is necessary to identify the illicit Bitcoin transactions as well as the normal transactions accurately. Since there are only 108 anomalous cases in this dataset, the under-sampling techniques will select the same number of non-anomalous cases. Also, the models will learn the positive samples that help to classify the anomalous transactions from the independent test set more correctly. Moreover, the over-sampling methods generate artificial synthetic data to equalize the minority and the majority samples \cite{ir18}. The synthetic oversampled data minimizes the false positive rates in the test dataset, however, can't perform well for highly imbalanced data. For this, variations of Generative Adversarial Network (GAN) based over-sampling techniques have been investigated by researchers for generating artificial data to balance the minority and majority class \cite{ir14}. Although there are several under-sampling algorithms, no technique is free from limitations. A major problem in most of the under-sampling algorithms is that significant instances which have a great impact on model training may be missed \cite{ir13}. Besides, overfitting problems, sensitivity to noises, introducing bias into the database, etc. can reduce the performance of the machine learning models \cite{ir17}. Still, there is no perfect under-sampling technique proposed as the dataset varies. Considering these issues, we have proposed an under-sampling technique based on the ensemble method. Our proposed algorithm selects a subset of instances from the majority class which has a significant impact on the performance of the machine learning models. Our algorithm gives importance to all the samples in the dataset by iteratively selecting the subsets and this minimizes the chance to miss the significant samples. Also, some combined balancing techniques have been investigated to compare which technique performs better in Bitcoin anomaly detection. 

After balancing the data, selecting a machine learning classifier is another challenging task. Tree-based machine learning classifiers have been used in many studies to classify malicious activities \cite{ir9} since faster training can be performed on tree-based classifiers \cite{ir10}. However, several studies show that the ensemble method can perform better in the case of large-scale data e.g. Bitcoin transactions than a single machine learning algorithm \cite{ir11}. The idea behind the stacked ensemble method is that by combining multiple models, the strengths of each model can be leveraged while mitigating their weaknesses \cite{ir12}. The stacked ensemble method differs from other ensemble methods, such as bagging and boosting, in that it combines models with different algorithms and/or hyperparameters rather than replicating the same model multiple times. This allows for greater model diversity and can lead to better performance. On the other side, Voting ensemble models predict the output by using the highest majority voting (Hard Voting) or the highest average value of the individual prediction (Soft Voting) \cite{ir19}. Although the complexity and computational costs may increase due to training multiple models in the base classifier, there are some valid reasons for using the stacking ensemble method. Firstly, Tree-based models are computationally faster than other Machine Learning (ML) models like SVM or KNN. Secondly, tree-based models can handle the data without any conversion or normalization \cite{r29}. Thirdly, A stacking-based ensemble model can increase performance by minimizing the prediction errors caused by the variance components. Moreover, the Voting Classifier with only tree-based models is a powerful and interpretable ensemble learning technique. Combining the predictions of different tree-based algorithms can provide more accurate and robust predictions, making it an attractive choice for various classification and regression tasks. Additionally, the interpretability of tree-based models remains preserved in the ensemble, which can be a valuable asset in applications where model transparency and understanding are essential. Considering the effectiveness of the combined tree-based methods, both stacked and voting ensemble models are proposed in this study.

After performing all these great tasks for anomaly detection, a question can arise "Should we trust the prediction of the Black-Box model?". XAI, Explainable Artificial Intelligence, is a field of interest to find the answer to this question. This latest AI technique helps to increase the explainability and transparency of the black-box AI models by making complex interpretable decisions \cite{ir20}. Two popular XAI techniques e.g. ‘Local Interpretable Model-Agnostic Explanations’ (LIME) \cite{ir21} and ‘SHapely Additive exPlanations’ (SHAP) \cite{ir22} have been used by researchers to prove the explainability and transparency of the AI models. SHAP has been investigated in our study since it can find interpretable decisions more quickly and accurately.         

In this study, an eXtreme Gradient Boosting-based Clustering (XGBCLUS) under-sampling method has been proposed and compared to other state-of-the-art under-sampling techniques to detect Bitcoin anomalous transactions. Besides, Two popular over-sampling techniques e.g. SMOTE and ADASYN have been used to generate synthetic data points for balancing majority and minority classes and also compared to the combined sampling methods SMOTE with Edited Nearest Neighbor (ENN) and SMOTE with TOMEK Links. A comparative analysis has been performed among the down-sampling and over-sampling methods to decide which sampling technique is appropriate for identifying anomalous transactions. To classify non-illicit and illicit Bitcoin transactions, tree-based ML classifiers such as Extreme Gradient Boosting (XGB) \cite{r23}, Random Forest (RF) \cite{r24}, Decision Tree (DT) \cite{r25}, Gradient Boosting (GB) \cite{r26}, and Adaptive Boosting (AdB) \cite{r27} have been used. Although a single ML classifier may work well for anomaly detection on the training dataset, however, the final prediction on the independent test dataset may perform poorly. That's why both stacking and voting ensemble models have been proposed to increase the accuracy by combining the outcomes from individual classifiers. The main reason behind this is if one of the chosen ML classifiers in the ensemble does not perform well, the risk can be minimized by averaging the outputs of all of them. Cross-validation has been used to avoid overfitting where the beginning training dataset is used to create multiple mini-train-test splits. 10-fold cross-validation has been used by partitioning the data into 10 subsets known as the fold. Then, iteratively the algorithm has been trained on 9 folds while the remaining fold is kept as the test set. Finally, the correctness of the prediction by the Black-Box ML models has been proved by the XAI technique i.e. using SHAP analysis. A set of rules is also presented to conduct interpretability analysis for determining whether a Bitcoin transaction is anomalous or not. The contributions of this study are summarised below.
\begin{itemize}
  \item We introduce an under-sampling algorithm based on eXtreme Gradient Boosting (XGBoost) called XGBCLUS, and we compare it with state-of-the-art methods.   
  \item We also explore various over-sampling and combined sampling techniques for classifying Bitcoin transactions. 
  \item Further, we compare the effectiveness of both under-sampling and over-sampling techniques, and we also compare the tree-based ensemble classifiers with the individual ML classifiers for anomaly detection.
  \item We explain the predictions of the ensemble models using SHAP (an eXplainable Artificial Intelligence technique) and identify the crucial features that exert the most influence on classifying Bitcoin transactions. 
  \item Lastly, we present a set of rules derived from a tree-based model to conduct interpretability analysis for anomalous transactions. 
\end{itemize}

The rest of the paper is structured as follows: Section \ref{rw} provides a summary of recent research involving both supervised and unsupervised machine learning techniques. In Section \ref{method}, we outline the specifics of the proposed methods. Comparative results are presented in Section \ref{ra}. Section \ref{disc} contains a discussion, and, lastly, Section \ref{cons} concludes the research paper.

\section{Related Work}\label{rw}
Researchers have paid attention to detecting or predicting whether a transaction is anomalous or not using the concept of blockchain intelligence \cite{r1}, i.e., introducing Artificial Intelligence (AI) for anomaly detection or fraud detection in blockchain transaction data. Several supervised and unsupervised techniques along with various balancing methods have been applied to detect fraudulent transactions in blockchain networks.
\subsection{Supervised Techniques}\label{st}
Chen et al. \cite{r9} employed supervised machine learning classifiers, including Random Forest (RF), Adaptive Boosting, MLP, SVM, and KNN, to detect bitcoin theft. RF exhibited the best performance with an F1 value of 0.952, surpassing other unsupervised algorithms. Another study \cite{r2} explored the Bitcoin ecosystem, categorizing illegal activities and using Bagging and Gradient Boosting for classification. While visualizing results, a notable limitation was the absence of proper balancing techniques for model tuning. Singh et al. \cite{r5} utilized SVM, Decision Tree, and Random Forest for Ethereum network anomaly detection, detecting anomalous transactions with some limitations in dataset coverage. Active learning tools were applied in a study \cite{r7} for illegal activity detection in Bitcoin transactions, claiming superiority over unsupervised methods. A comparative study \cite{r18} favored ensemble-based methods for classifying non-anomalous and anomalous Bitcoin transactions based on accuracy and F1-score metrics.  
\subsection{Unsupervised Techniques}\label{ust}
In their analysis \cite{r6}, the authors used Bitcoin transaction data to create two graphs for users and transactions, aiming to detect anomalies. Employing unsupervised methods such as SVM, K-means clustering, and Mahalanobis distance, they identified two anomalous users and one anomalous transaction out of 30 cases. However, the method struggled to identify maximum positive cases. Another study \cite{r3} utilized the K-means algorithm for anomaly detection in blockchain electronic transactions, also incorporating OSVM to find outliers. This approach suffered from high false positive rates. In a comparative study \cite{r4}, various unsupervised learning algorithms, including IForest, One Class SVM, Two Phase Clustering, and Multivariate Gaussian, were evaluated, with the Multivariate Gaussian algorithm showing the highest F1-Score. Evaluating the trimmed K-Means clustering algorithm, Monamo et al. \cite{r8} successfully identified 5 anomalous activities out of 30 cases related to illicit transactions in the Bitcoin network. An encoder-decoder-based deep learning model was designed to detect anomalous activities in the Ethereum transaction network, claiming to detect illicit activities in the Ethereum network for the first time \cite{r11}. Lastly, clustering and role detection methods were applied in a study \cite{r10} to identify suspicious users in Bitcoin transaction data, using K-means for clustering and RoIX for role detection.  
\subsection{Balancing Techniques}\label{bt}
Researchers, as highlighted in \cite{r30}, have proposed various undersampling and oversampling methods to enhance evaluation metrics. For instance, in a comparative study \cite{r13}, customized nearest-neighbor undersampling achieved 99\% accuracy, outperforming various SMOTE-based oversampling techniques in analyzing Bitcoin and Ethereum transaction data. In the realm of bank transactions, ensemble-based classifiers, particularly SVM SMOTE dataset balancing with the Random Forest classifier, were found effective in detecting fraudulent activities \cite{r14}. Similarly, authors in \cite{r15} proposed an undersampling technique using fuzzy C-means clustering and similarity checks for identifying illicit activities in credit card transactions. Additionally, SMOTE oversampling has been employed in studies such as \cite{r16} and \cite{r17} to classify illegal activities in credit card transactions, while various undersampling techniques were investigated in \cite{r19} and \cite{r20} to discern normal and illegal activities. Notably, for handling highly imbalanced datasets, \cite{ahmed2022comparative} explored several undersampling techniques for early product back-order prediction. 
\section{Methodology}\label{method}
\begin{figure*}[h]
    \centering
    \includegraphics[width=0.9\textwidth]{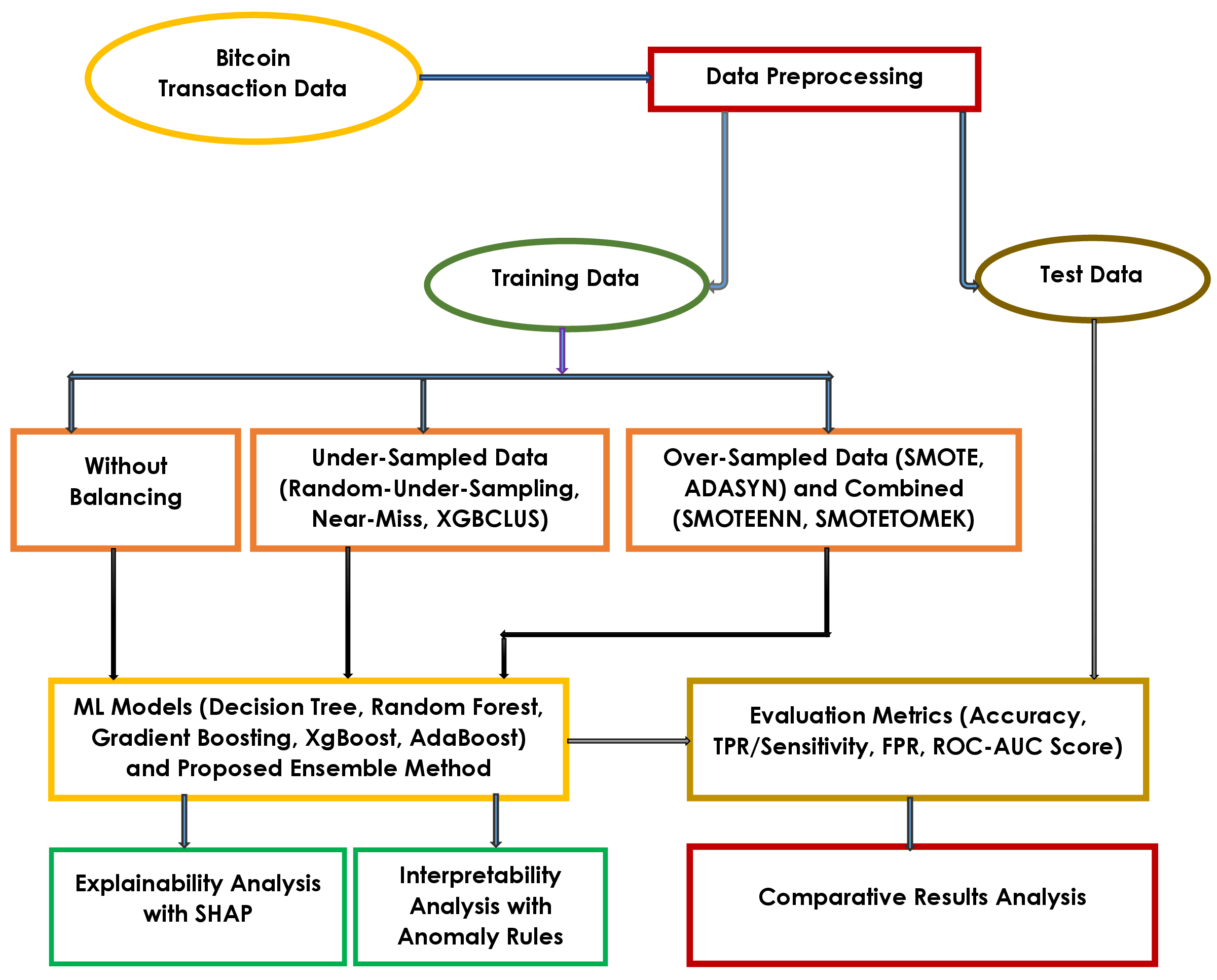}
    \caption{Methodology of this study}
    \label{fig1}
\end{figure*}

The overall methodology of this study is illustrated in Figure \ref{fig1}. Initially, we collected the dataset and conducted necessary preprocessing. Subsequently, the dataset was divided into training and testing data. Data sampling was exclusively applied to the training data, while the test data remained independent. The sampled data was employed to train both individual and ensemble machine learning classifiers. Finally, the independent test data was utilized to validate the models, employing various evaluation metrics including Accuracy, TPR, FPR, and ROC-AUC score. Additionally, we conducted a comparative analysis, combined with eXplainable Artificial Intelligence (XAI) and rules from Decision Tree, which is discussed in Section \ref{ra}.
\subsection{Dataset}\label{ds}
\begin{figure*}[h]
    \centering
    \includegraphics[width=0.9\textwidth]{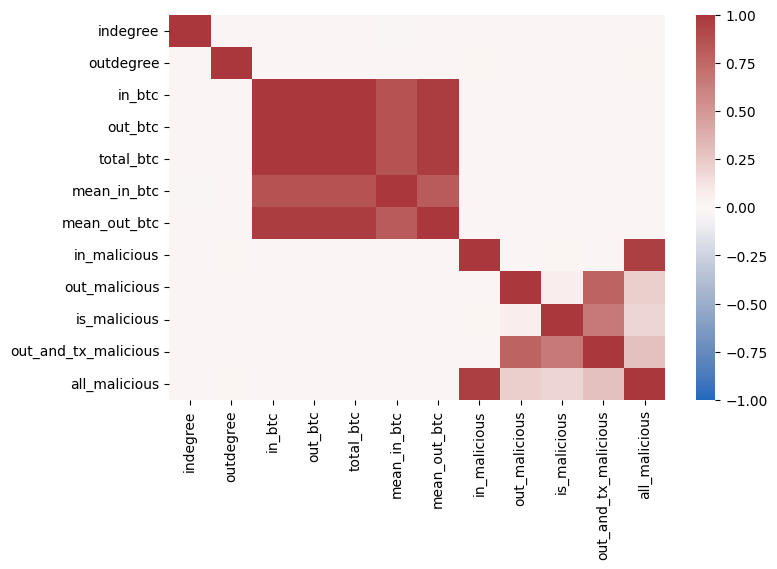}
    \caption{Correlations among the features using Heatmap}
    \label{fig111}
\end{figure*}

The bitcoin transaction data has been collected from the IEEE Data Portal\footnotemark[1]\footnotetext[1]{https://ieee-dataport.org/open-access/bitcoin-transactions-data-2011-2013}. It contains a total of 30,248,134 samples where 30,248,026 has been labeled as negative samples i.e. the non-malicious transactions. On the other hand, there are only 108 malicious samples. So it is clear that the dataset is highly imbalanced. Exploratory Data Analysis (EDA) has been performed to reveal the dataset insights. There exist 12 attributes with a label to indicate whether a Bitcoin transaction is anomalous or not. The anomalous transaction is labeled by 1 and 0 stands for non-anomalous transactions. 
\begin{table}[H]
  \caption{The T-values and P-values for all attributes}
    \centering
    \begin{tabular}{c c c}
    \hline
        Attribute & t value & p value \\\hline
        indegree & -14.013838 & 0.000000\\
        outdegree & 0.842249 & 0.399648\\
        in\_btc & -17.229753 & 0.000000\\
        out\_btc & -16.469202 & 0.000000\\
        total\_btc & -16.864202 & 0.000000\\
        mean\_in\_btc & -8.727102 & 0.000000\\
        mean\_out\_btc & -16.014732 & 0.000000\\
        in\_malicious & -68.869826 & 0.000000\\
        out\_malicious & -5432.702805 & 0.000000\\
        is\_malicious & -3878.622465 & 0.000000\\
        all\_malicious & -1866.899584 & 0.000000\\
\hline
    \end{tabular}

    \label{tab:tvalue}
\end{table}
A correlation matrix of the features is depicted in Figure \ref{fig111}. Features such as \textit{in\_btc, out\_btc, total\_btc, mean\_in\_btc,} and \textit{mean\_out\_btc} exhibit strong positive correlations. Conversely, the \textit{indegree} and \textit{outdegree} features display weak correlation. While \textit{in\_malicious, out\_malicious, is\_malicious,} and \textit{all\_malicious} features are notably correlated with the output feature \textit{out\_and\_tx\_malicious}, no substantial correlation is observed between these features and \textit{indegree, outdegree, in\_btc, out\_btc, total\_btc, mean\_in\_btc,} and \textit{mean\_out\_btc}.

Furthermore, we conducted a hypothesis test for feature selection employing the T-test to examine correlations between positive (anomalous Bitcoin transactions) and negative (non-anomalous Bitcoin transactions) samples. The T-test determines whether a significant difference exists between the means of the positive and negative samples. T-statistic values and corresponding p-values are computed for each attribute. These values are presented in Table \ref{tab:tvalue}. Notably, all attributes demonstrate significance with p-values below 0.01, except for \textit{outdegree}. Additionally, it's worth noting that all features (\textit{in\_malicious, out\_malicious, is\_malicious,} and \textit{all\_malicious}) share the same value range as the target feature \textit{out\_and\_tx\_malicious}. Consequently, this similarity poses a challenge for ML classifiers in effectively discerning Bitcoin transactions, leading to the exclusion of these four features and \textit{outdegree}. Ultimately, a set of seven features, including the target feature, is selected for classification. A summary of the selected features is shown in Table \ref{tab1}.
\begin{table*}[h]
\caption{Summary of the selected features}\label{tab1}%
\begin{tabular*}{\linewidth}{l @{\extracolsep{\fill}}c c}
\toprule
Feature Name & description\\
\midrule
Indegree    & No. of inputs for a given transaction \\[2pt]
in\_btc    & No. of bitcoins on each incoming edge to a given transaction \\[2pt]
out\_btc    & No. of bitcoins on each outgoing edge from a given transaction \\[2pt]
total\_btc    & Total number of bitcoins for a given transaction\\[2pt]
mean\_in\_btc    & Average number of bitcoins on each incoming edge to a given transaction \\[2pt]
mean\_out\_btc    & Average number of bitcoins on each outgoing edge from a given transaction \\[2pt]
out\_and\_tx\_malicious    & Status of a given transaction if it is malicious or not \\[2pt]
 \bottomrule
\end{tabular*}
\end{table*}
\begin{figure}[h]%
    \centering
    \includegraphics[width=0.45\textwidth]{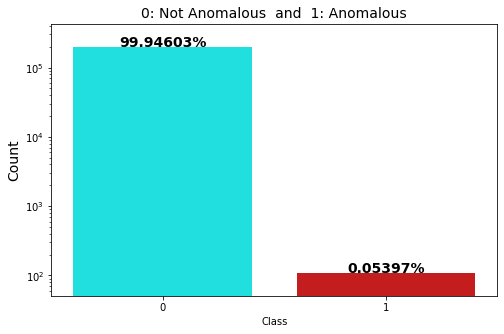}
    \caption{Class ratio}
    \label{fig2}
\end{figure}

 Following the feature selection process, the negative and positive samples were segregated, and duplicate entries were eliminated exclusively from the negative samples. To mitigate computational complexity, we opted to retain only 200,000 negative samples, accompanied by 108 positive samples. Nevertheless, the resulting imbalance ratio remains considerably high as shown in Figure \ref{fig2}. 
\subsection{Imbalanced Data Handling}\label{idh}
In Bitcoin transactions, the number of illicit transactions is significantly lower than that of normal transactions, leading to an imbalanced dataset. Consequently, machine learning classifiers tend to exhibit bias toward the majority class \cite{r21}. While classification accuracy might appear satisfactory in many cases, a notable discrepancy between TPR and FPR values often arises—indicating that the models struggle to accurately classify anomalies. To address this, it is crucial to balance the dataset using built-in or customized sampling techniques prior to training the classification models.
In scenarios involving anomaly detection, fraud identification, or money laundering, positive cases are typically scarce. As such, undersampling techniques can prove effective in rebalancing the dataset while prioritizing accurate identification of positive cases. However, in instances where the number of positive cases or anomalies in the minority class is exceedingly low, ML classifiers may be trained on a limited dataset generated by the undersampling technique.
On the other hand, over-sampling methods aim to increase the instance count of the minority class to match that of the majority class. Despite generating artificial data based on a combination of majority and minority samples, these methods can be effective. In our study, we introduce an under-sampling algorithm named XGBCLUS, and we also investigate established under-sampling techniques such as Random Under Sampling (RUS) and Near-Miss. Furthermore, we explore popular over-sampling techniques including SMOTE, ADASYN, as well as combined approaches like SMOTEENN and SMOTETOMEK. We present a comparison between over-sampling and under-sampling methods in Section \ref{ra}.
\subsubsection{Under-Sampling Techniques}
We have investigated two under-sampling methods e.g. Random Under Sampling and Near-Miss along with our proposed under-sampling method which is described in Section \ref{proposed algorithm}. Random Under Sampling (RUS) is a simple technique used to handle class imbalance in datasets. It randomly selects a subset of instances from the majority class. The size of this subset is determined based on the desired balance ratio between the minority and majority classes. The balance ratio $\alpha_u{}_s$ is defined by Equation \ref{balance ratio}.
\begin{equation}
    \alpha_u{}_s = \frac{N_m}{N_r{}_m}
    \label{balance ratio}
\end{equation}
where $N_m$ is the number of samples in the minority class and $N_r{}_m$ is the number of samples in the majority class after resampling. Then, the instances from both the minority class and the randomly selected subset of the majority class instances are combined to form a balanced dataset. Another under-sampling technique namely Near-Miss is also used for balancing the dataset. It reduces the imbalance by retaining a subset of instances from the majority class that are close to instances from the minority class. This down-sampling technique selects instances based on their proximity to the minority class, making it possible to preserve important samples. For each instance in the minority class, Near-Miss calculates its distances to all instances in the majority class using various distance metrics such as Euclidean distance or Manhattan distance. After that, this algorithm identifies the k instances from the majority class that are closest to each instance in the minority class. The value of k is typically set as a hyperparameter and determines the degree of under-sampling. We set the value as 1 and hence we call it Near-Miss-1. Finally, it combines the instances from both the minority class and the selected instances from the majority class to form an under-sampled dataset. 
\subsubsection{XGBCLUS Algorithm}\label{proposed algorithm}
XGBCLUS (eXtreme Gradient Boosting-based Clustering) operates by merging clusters ( accomplished by selecting random instances from the majority class in the training data, equal in number to the positive instances from the minority class) and the Extreme gradient Boosting algorithm. 
\begin{algorithm}
\caption{XGBCLUS algorithm}
\label{alg:xgbclus}
\begin{algorithmic}[1] 
\State \textbf{Input:} Imbalanced Training samples, \textit{DATA}; Number of iterations, \textit{k}; Number of positive samples,\textit{P}; The independent test data and The XGBoost algorithm.
\State \textbf{Output:} Selected Under-Sampled data
    
    \State Initialize $TMAX$ and $FMIN$ 
    \State Initialize an empty set \textit{Selected\_Samples}
    \For{$i = 1$ to $k$}
        \State Select $n$ negative samples arbitrary equal to $P$ and Prepare the Train data
        \State Train the model and predict using the test samples 
        \State Calculate True Positive (TP) and False Positive (FP) values
        \If{$TP > TMAX$ and $FP < FMIN$ }

            \State Set $TMAX = TP$ and $FMIN = FP$ 
            \State Update current $n$ samples in \textit{Selected\_Samples}  
        \EndIf 
    \EndFor
    \If{\textit{Selected\_Samples} is empty}
    \State \textbf{Goto} step $3$ and repeat the steps $3$ - $13$ after changing $TMAX$ and $FMIN$ values
    \Else
    \State \textbf{Return }$Selected\_Samples$
    \EndIf
\end{algorithmic}
\end{algorithm}
The algorithm starts with splitting the whole dataset into train and test data. The test data is kept independent and the positive samples, \textit{\textbf{P}}, are counted from the training set. Then, the number of iterations,  \textit{\textbf{k}}, is calculated by dividing the total number of negative samples in the training data by the positive samples,  \textit{\textbf{P}}. 

In each iteration, the algorithm arbitrarily selects \textit{\textbf{n}} negative samples equal to  \textit{\textbf{P}} and a new training set is prepared to train the Xgboost model. Using the independent test set, the model predicts the True Positive (TP) and False Positive (FP) values. The values of TP and FP values are compared with TMAX and FMIN, respectively. TMAX and FMIN are initialized with arbitrary values where the TMAX represents the maximum true positive value and the minimum false positive value is defined by FMIN. If the TP value is greater than the TMAX value and the value of FP is less than the FMIN value, then both TMAX and FMIN are updated with the TP and FP values respectively. At the same time, current \textit{\textbf{n}} samples are updated in the \textit{Selected\_Samples} set. Otherwise, no changes are made in the current iteration.

After the \textit{\textbf{k}} iterations are finished, the \textit{Selected\_Samples} set is checked. If the set is empty, the algorithm should be run again with new TMAX and FMIN values. Otherwise, the samples in the \textit{Selected\_Samples} set are the under-sampled data returned by the algorithm. The XGBClus algorithm is shown in Algorithm \ref{alg:xgbclus}. 
\subsubsection{Over-Sampling Techniques}
We have also investigated two popular over-sampling and two combined sampling strategies for handling the class imbalance in Bitcoin transaction data. Among them, Synthetic Minority Over-sampling Technique (SMOTE) is a widely used technique to handle the class imbalance problem, especially in Bitcoin Transactions for detecting anomalies. It aims to balance the class distribution by generating synthetic examples of the minority class, thereby mitigating the bias and improving model generalization. For each minority sample $X_i$, it selects the k nearest neighbors randomly from the same class. A new sample $X_n$ is generated using one of the nearest neighbors $X_z{}_i$ from k and the new sample is generated using the equation \ref{new sample}.  
\begin{equation}
    {X_n} = {X_i}+\lambda * ({X_z{}_i} - {X_i})   
    \label{new sample}
\end{equation}
where $\lambda$ is a random number between 0 and 1. Thus, a new synthetic instance is generated in the feature space. This process is repeated until the samples in both majority and minority classes are the same. At last, the original minority instances are combined with the newly generated synthetic instances to form a balanced dataset. 

ADASYN (Adaptive Synthetic Sampling) also uses the same formula for generating the new sample except for the selection of $X_i$. It increases the density of synthetic instances in the regions that are harder to classify which provides a more refined way of handling class imbalance. For each minority instance, it first calculates the number of its k nearest neighbors that belong to the majority class to get an idea of how close the minority sample is to the majority class. Then, it computes the imbalance ratio $\alpha_o{}_s$ for each minority instance using the equation \ref{imbalance ratio}.
\begin{equation}
    \alpha_o{}_s = \frac{N_r{}_m}{N_m}
    \label{imbalance ratio}
\end{equation}
where $N_r{}_m$ is the number of samples in the minority class after resampling and $N_m$ is the number of samples in the majority class. This imbalance ratio is used to calculate the desired number of synthetic instances to be generated for the current minority instance. A difficulty ratio is also calculated to discover the hard level. If the ratio is high, it considers more neighbors for generating synthetic instances. By interpolating feature values between the minority instance and its selected neighbors, a new synthetic instance is generated. This process is repeated for all minority instances until the samples in both majority and minority classes are the same. Finally, the original minority instances are combined with the newly generated synthetic instances to form a balanced dataset. 

SMOTEENN is a combination of two resampling techniques, SMOTE (Synthetic Minority Over-sampling Technique) and Edited Nearest Neighbors (ENN). It tackles class imbalance by creating synthetic samples using SMOTE and subsequently refining the dataset using ENN to eliminate potentially noisy or misclassified instances. SMOTEENN aims to provide a more refined approach to balancing imbalanced datasets while also improving the quality of the final dataset by eliminating potential noise. After generating the synthetic instances by SMOTE, for each instance in the dataset, ENN first finds the k nearest neighbors. If the instance's class is different from the majority class of its neighbors, it removes the instance from the dataset. Thus, ENN eliminates noisy or misclassified instances. Another combined resampling technique is SMOTETOMEK where synthetic instances are generated by SMOTE and the under-sampling technique TOMEK Link identifies pairs of instances from different classes that are closest to each other. For each pair of instances identified as Tomek links, it removes the instance from the majority class. This undersampling method helps in removing instances that are close to the decision boundary and might be misclassified. This can lead to a more balanced, discriminative, and effective dataset for training machine learning models. 
\subsection{Proposed Ensemble Model}\label{em}
The meta-classification ensemble method based on stacked generalization is a Machine Learning (ML) approach used to improve the accuracy of predictions by combining multiple models \cite{r22}. The stacking-based ensemble model is formed by two classifiers. One is the base classifier and the other is the meta classifier. It starts with training a set of base models using several classifiers. A new dataset is found from the base-level classifiers and then the meta-classifier, also known as a combiner or a blender, is trained using the new dataset. After that, the learned meta-classifier is used to predict the independent test dataset. The architecture of the proposed stacked-ensemble model is shown in Figure 4. 
\begin{figure*}[h]%
    \centering
    \includegraphics[width=0.9\textwidth]{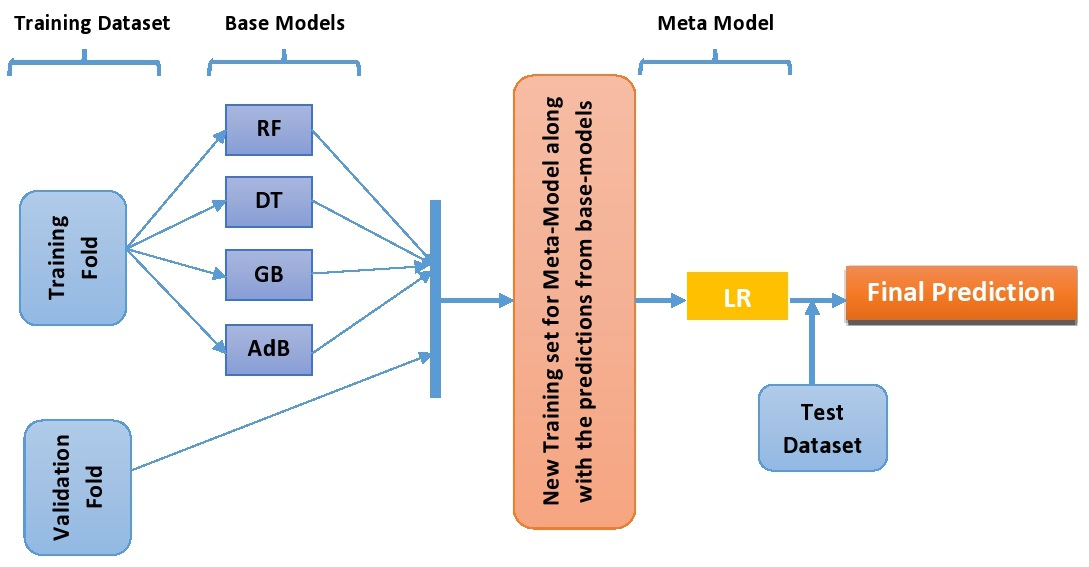}
    \caption{Stacked-Ensemble Model architecture}
    \label{fig444}
\end{figure*}

In our proposed stacking-based ensemble model, Random Forest (RF), Decision Tree (DT), Gradient Boosting (GB), and Adaptive Boosting (AdB) have been used as the base models. The training dataset is used to train the base models and the outputs of the four base models along with the validation fold are combined to create a new dataset. Then Logistic regression (LR) \cite{r28}, which is the meta-classifier in our proposed model, receives the newly formed dataset by combining the predictions of the base classifiers as input and learns on that input set. Finally, the test dataset has been used to predict anomalous and non-anomalous transactions.

On the other side,  the Voting Classifier is constructed using only tree-based models, which are a family of machine learning algorithms known for their robustness and interpretability. The architecture of a Voting Classifier that incorporates tree-based models like Decision Trees (DT), XGBoost (XGB), Gradient Boosting (GB), Random Forest (RF), and AdaBoost (ADB) is shown in Figure 5. The Voting Classifier takes the training set as input and the ensemble is constructed by combining the predictions of several individual tree-based models. Each model in this context refers to a unique instantiation of the tree-based algorithm with a specific set of hyperparameters or configurations. After the Voting Classifier has been trained and evaluated, it is used to make predictions on test data. The Voting Classifier aggregates the predictions of each individual tree-based model using a voting mechanism. The voting can be either "hard" or "soft". In hard voting, each model in the ensemble casts a single vote for the predicted class label, and the majority class receives the final prediction. For soft voting, the probabilities (confidence scores) of each model's predicted classes are averaged, and the class with the highest average probability is chosen as the final prediction.  

\begin{figure*}[h]%
    \centering
    \includegraphics[width=0.9\textwidth]{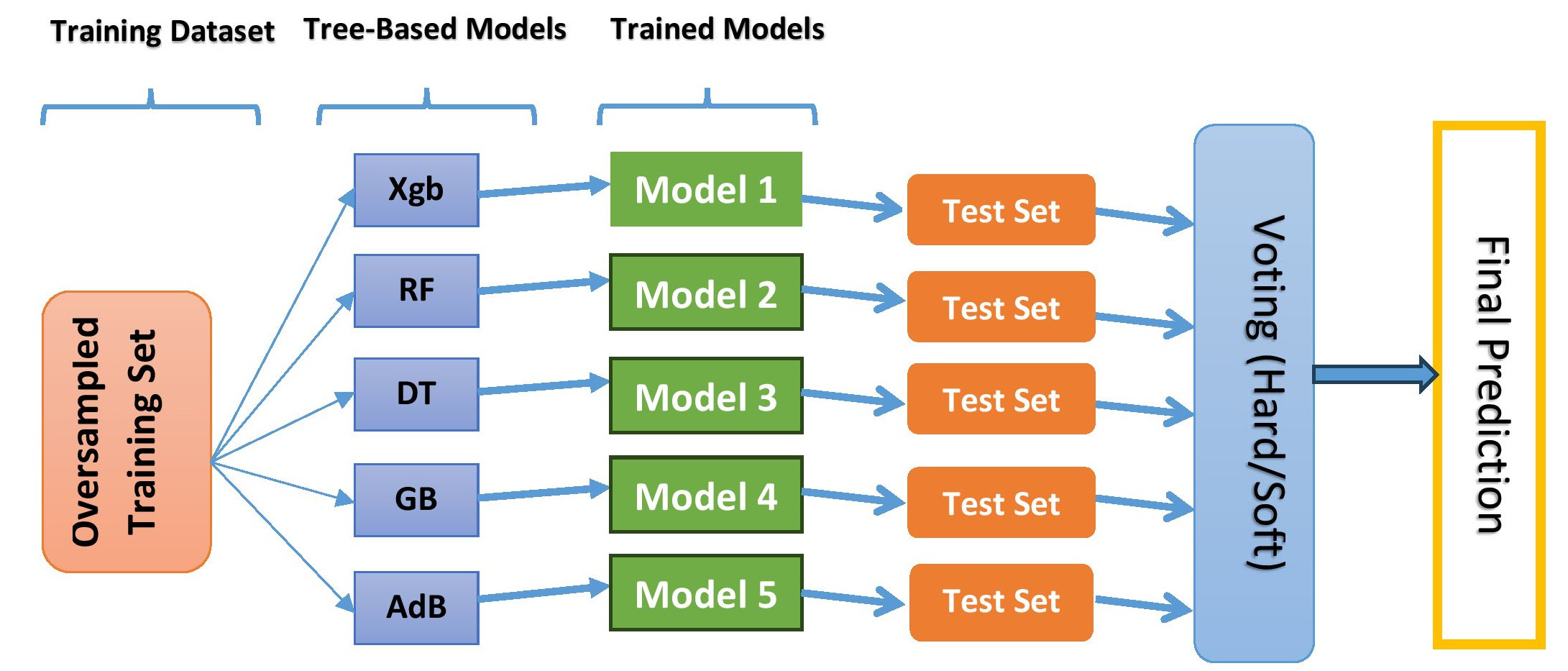}
    \caption{Voting-Ensemble Model architecture}
    \label{fig5}
\end{figure*}

\section{Results Analysis}\label{ra}
In this section, we assess the performance of the proposed ensemble models using both under-sampled and over-sampled data. Additionally, a comprehensive comparative analysis between single classifiers and ensemble classifiers is presented herein. We commence by configuring the experimental environment and subsequently present outputs that scrutinize various facets of the model's performance concerning Bitcoin anomaly detection. Thorough performance analyses have been conducted on a dataset encompassing both normal and anomaly transactions within the realm of Bitcoin transactions. An overview of the dataset is provided in Section 3. To facilitate performance evaluation, several Python libraries such as Pandas, NumPy, and scikit-learn were employed. The Python script was executed on Colab Pro, utilizing 12.68 GB of RAM and 225.83 GB of disk space.
\subsection{Evaluation Metrics}
Given that accuracy alone is insufficient to gauge the performance of an anomaly detection system, it becomes crucial to employ additional metrics such as True Positive Rate (TPR) to assess the accurate identification of anomalous transactions and False Positive Rate (FPR) to evaluate the correct identification of non-anomalous transactions. This aligns with the primary objective of our study. The evaluation metrics e.g. accuracy, True Positive Rate (TPR), and False Positive Rate (FPR) have been used to compare the performance of the single models without and with balancing the data against the proposed ensemble models. Additionally, we also use the feature importance score to show the hierarchy of the features. We have also considered the Receiver Operating Characteristic (ROC) score, which compares the True Positive Rate (TPR) against the False Positive Rate (FPR).
The performance metrics are defined below: \\
\textit{TP} = True Positive: an anomalous transaction is correctly identified as anomalous \\
\textit{TN} = True Negative: a non-anomalous or normal transaction is correctly identified as non-anomalous\\
\textit{FP} = False Positive: a non-anomalous transaction is incorrectly identified as anomalous \\
\textit{FN} = False Negative: an anomalous transaction is incorrectly identified as non-anomalous
\begin{equation}
    Accuracy = \frac{TP+TN}{TP+TN+FP+FN}
\end{equation}
\begin{equation}
    TPR = Sensitivity = \frac{TP}{TP+FN}
\end{equation}
\begin{equation}
    TNR = Specificity = \frac{TN}{TN+FP}
\end{equation}
\begin{equation}
    FPR = \frac{FP}{TN+FP}
\end{equation}

AUC is calculated as the Area Under the $Sensitivity(TPR)$ - ($1-Specificity)(FPR)$ Curve.
\subsection{Effects of Under-Sampling in Classification}\label{eofus}
We have kept 20\% data for the independent test set and the remaining 80\% has been used for training the models. To prove the data imbalanced problem, the classifiers have been trained without balancing the train set. The ML classifiers become biased to the majority samples and result in a high true negative value. However, the ML classifiers can not identify the positive samples correctly that's why the true positive rate is very low and in some cases, it is zero. Table \ref{tab222} shows the comparison of accuracy, True Positive (TP), and roc-auc score of the Decision Tree (DT), Gradient Boosting (GBoost), Random Forest (RF), and Adaptive Boosting (AdaBoost) classifiers. The TP values are zero for DT and RF classifiers which indicates that no anomalous transactions are correctly identified and all transactions are classified as normal transactions. Although the accuracy seems to be good enough for the classifiers, the TP score tends to zero i.e. anomalous transactions are not identified because of the biases of models to the majority of transactions. Given that detecting anomalous transactions is the primary objective of our study, classifiers may struggle to identify those transactions without balanced data. Therefore, we explored several balancing techniques to enhance the True Positive Rate (TPR) and decrease the False Positive Rate (FPR) values.          
\begin{table*}[!ht]
\begin{center}

\caption{Comparison among the classifiers without balancing the data}\label{tab222}%
\begin{tabular*} {\linewidth}{l @{\extracolsep{\fill}}c c c c c c}
\toprule
Classifiers & Accuracy & TPR & AUC-Score\\
\midrule
DT & 0.99 & 0.0 & 0.55\\
GBoost & 0.99 & 0.09 & 0.62\\
RF & 0.99 & 0.0 & 0.72\\
AdaBoost & 0.99 & 0.05 & 0.82\\
\bottomrule
\end{tabular*}
\end{center}
\end{table*}

\begin{figure*}[h]
     \begin{subfigure}[b]{0.47\textwidth}
         \centering
         \includegraphics[width=3in,height=3in]{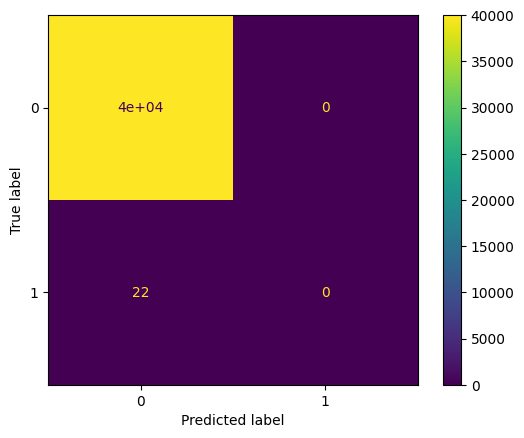}
         \caption{}
         \label{cmwithoutbalance}
     \end{subfigure}
     \begin{subfigure}[b]{0.47\textwidth}
         \centering
         \includegraphics[width=3in,height=3in]{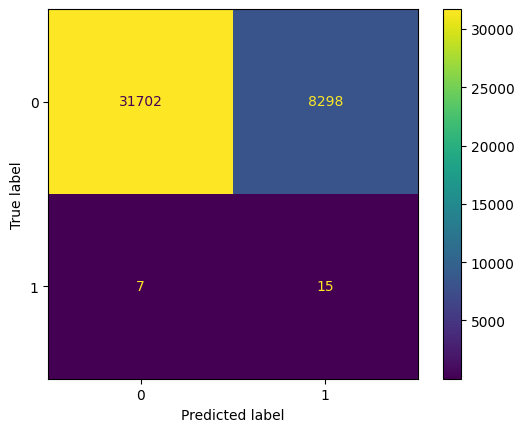}
         \caption{}
         \label{cmwithrandomsampling}
     \end{subfigure}
     \begin{subfigure}[b]{0.47\textwidth}
         \centering
         \includegraphics[width=3in,height=3in]{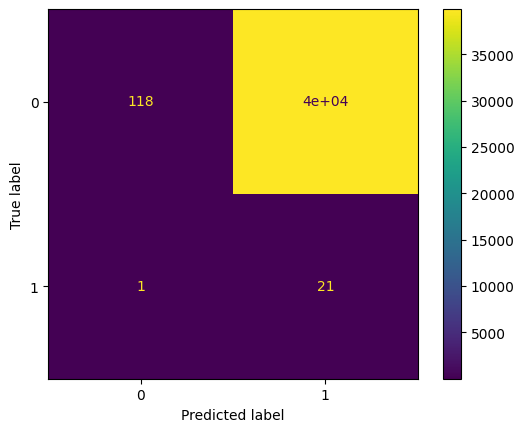}
         \caption{}
         \label{cmwithnearmiss}
     \end{subfigure}
     \begin{subfigure}[b]{0.47\textwidth}
         \centering
         \includegraphics[width=3in,height=3in]{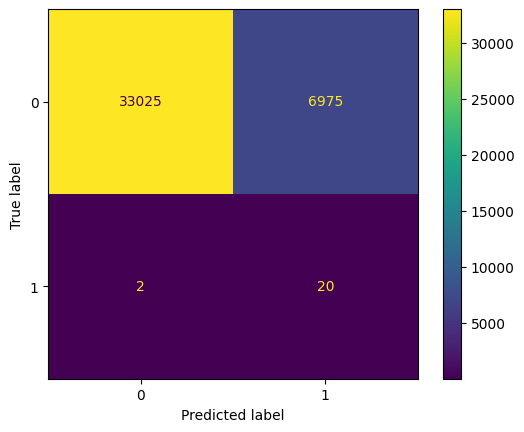}
         \caption{}
         \label{cmwithxgbclus}
     \end{subfigure}
    \caption{ Confusion Matrix for (a) Without Balancing, (b) Ensemble classifier with RUS, (c) Ensemble classifier with NearMiss1, (d) Ensemble classifier with XGBClus}
    \label{cm--with-under-sampling}
\end{figure*}

In Figure \ref{cm--with-under-sampling}, the confusion matrices illustrate the performance of the ensemble classifier under different under-sampling methods, namely Random Under Sampling (RUS), Nearmiss1, and XGBClus. Notably, the True Positive (TP) values exhibit an increase compared to scenarios without balancing, where TP values are consistently zero. However, it is essential to discern that, despite the improvements in TP, the False Positive (FP) values show variations among the under-sampling methods. Specifically, Nearmiss1 displays a relatively higher FP count compared to RUS and XGBClus, even though the FP is zero in the absence of balancing. The XGBClus undersampling method proposed in our study outperforms the existing method by achieving the highest True Positive (TP) value along with relatively low False Positive (FP) values. This superiority stems from our algorithm's approach of considering all instances in downsampling, whereas existing algorithms randomly select instances, leading to the omission of important cases.

\begin{figure*}[h]%
    \centering
    \includegraphics[width=5.5in,height=2.8in]{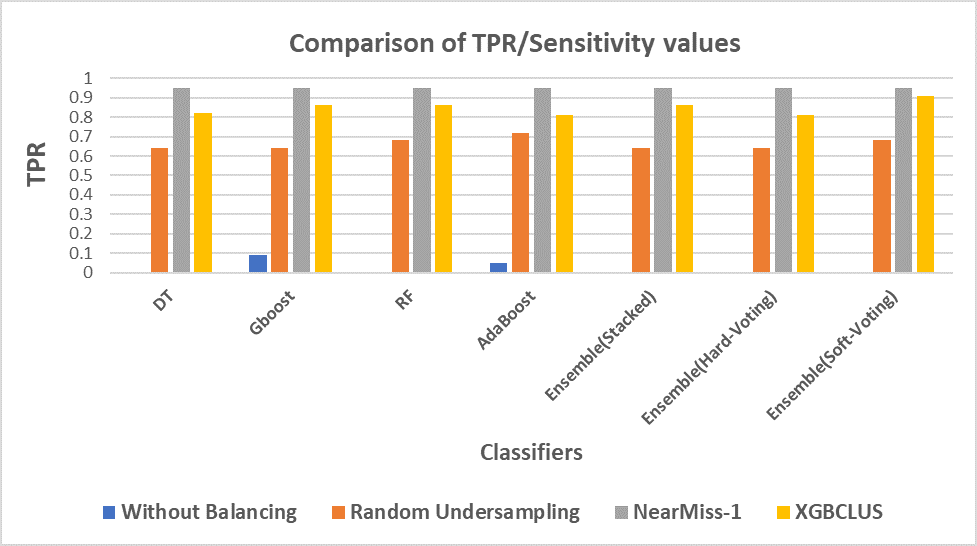}
    \caption{Comparison of TPR or sensitivity values using under-sampling methods}
    \label{fig6}
\end{figure*}

Given that the TPR or sensitivity signifies the count of correctly classified positive transactions, down-sampling techniques were explored to equalize the numbers of normal and anomalous transactions. We employed XGBCLUS, our proposed under-sampling method, in conjunction with other established techniques for downsampling. Figure \ref{fig6} illustrates that the sensitivity values for all single and ensemble ML classifiers witnessed an increase, except for the NearMiss-1 under-sampling algorithm. Notably, the sensitivity values for both single and ensemble classifiers utilizing the XGBCLUS algorithm stand at 0.82, 0.86, 0.86, 0.81, 0.86, 0.81, and 0.91, respectively. These values exceed the sensitivity values obtained without balancing and those from random under-sampling techniques. 

While the NearMiss-1 undersampling technique yields a higher sensitivity value compared to the XGBCLUS method, the corresponding FPR value is markedly high, as demonstrated in Table \ref{tabfpr}. As the False Positive Rate (FPR) decreases, the True Negative Rate (TNR) increases, indicating the correct identification of non-anomalous transactions. The NearMiss-1 undersampling technique exhibits a higher FPR, suggesting its limitation in accurately identifying non-anomalous transactions. In contrast, the random undersampling method produces average FPR values, although they are higher than the FPR values of XGBClus. In terms of TPR and FPR values, XGBClus outperforms other undersampling techniques.

\begin{table*}[ht]
\begin{center}

\caption{FPR of ML classifiers after Under-Sampling}\label{tabfpr}%
\begin{tabular*} {\linewidth}{l @{\extracolsep{\fill}} c c c c c}
\toprule
Classifiers & Random Undersampling & NearMiss 1 & XGBCLUS\\
\midrule
DT & 0.26 & 0.99 & 0.18\\
GBoost & 0.26 & 0.99 & 0.19\\
RF & 0.20 & 0.99 & 0.16\\
AdaBoost & 0.26 & 0.99 & 0.21\\
Ensemble (Stacked) & 0.22 & 0.99 & 0.15\\
Ensemble (Hard-Voting) & 0.21 & 0.99 & 0.14\\
Ensemble (Soft-Voting) & 0.21 & 0.99 & 0.17\\
\bottomrule
\end{tabular*}
\end{center}
\end{table*}

\begin{figure*}[h]%
    \centering
    \includegraphics[width=0.9\textwidth]{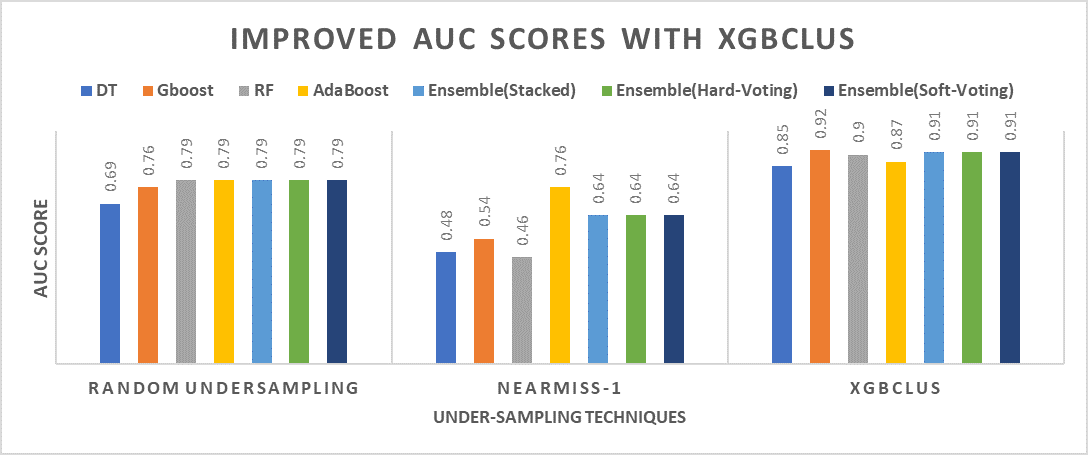}
    \caption{Improved ROC-AUC value using XGBClus}
    \label{fig7}
\end{figure*}

Figure \ref{fig7} illustrates the enhanced ROC-AUC scores obtained through the utilization of under-sampled data. Notably, the proposed XGBCLUS undersampling technique outperforms RUS and Near-Miss in terms of ROC-AUC scores. The Gradient Boosting (GBoost) classifier achieves the highest ROC-AUC score of 0.92, which stands as the peak among all single and ensemble classifiers. Furthermore, the remaining classifiers achieve ROC-AUC scores ranging from 0.85 to 0.91. This range indicates that the true positive and false positive rates adhere to their expected levels.

\subsection{Effects of Over-Sampling in Classification} \label{eofos}
\begin{figure*}[h]
     \begin{subfigure}[b]{0.47\textwidth}
         \centering
         \includegraphics[width=3in,height=3in]{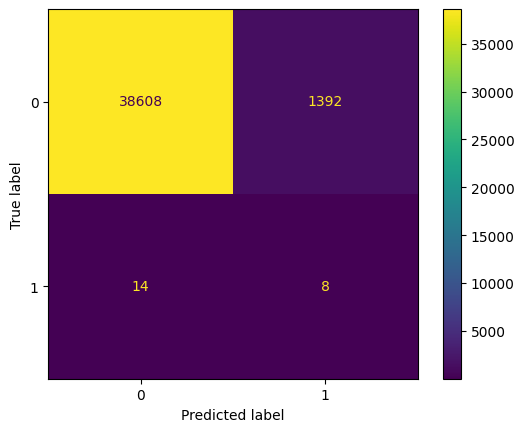}
         \caption{}
         \label{cmwithsmote}
     \end{subfigure}
     \begin{subfigure}[b]{0.47\textwidth}
         \centering
         \includegraphics[width=3in,height=3in]{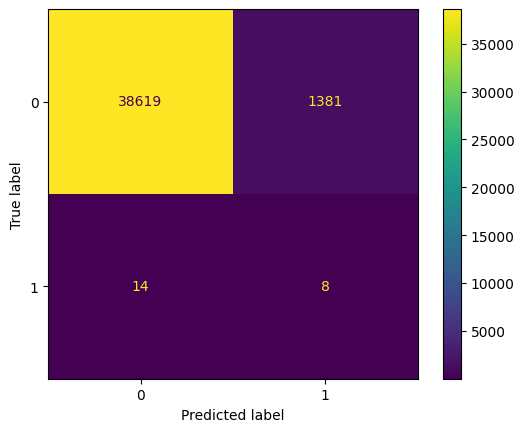}
         \caption{}
         \label{cmwithadasyn}
     \end{subfigure}
     \begin{subfigure}[b]{0.47\textwidth}
         \centering
         \includegraphics[width=3in,height=3in]{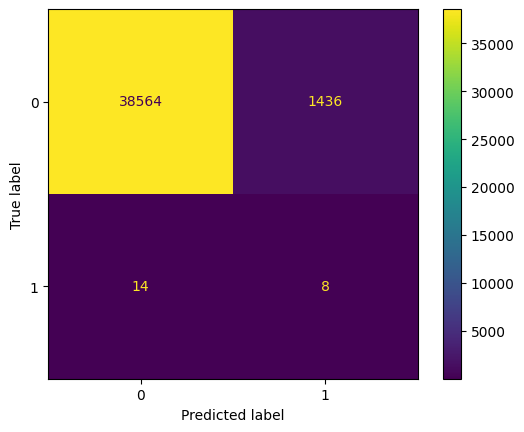}
         \caption{}
         \label{cmwithsmoteenn}
     \end{subfigure}
     \begin{subfigure}[b]{0.47\textwidth}
         \centering
         \includegraphics[width=3in,height=3in]{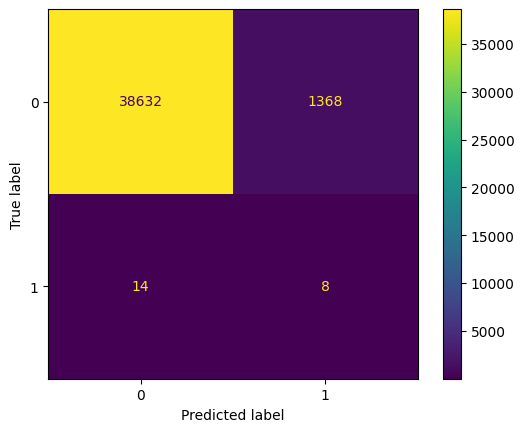}
         \caption{}
         \label{cmwithsmotetomek}
     \end{subfigure}
    \caption{ Confusion Matrix for (a) Ensemble classifier with SMOTE, (b) Ensemble classifier with ADASYN, (c) Ensemble classifier with SMOTEENN, (d) Ensemble classifier with SMOTETOMEK}
    \label{cm-with-over-sampling}
\end{figure*}

Although the TPR scores have increased for all ML classifiers following the under-sampling of data, the FPR values remain unsatisfactory. The objective is to elevate the True Positive (TP) rate while reducing the FP rate. To achieve this balance, various over-sampling and combined methods have been applied to balance the training data. In Figure \ref{cm-with-over-sampling}, the confusion matrices depict the performance of the ensemble classifier when employing various over-sampling methods, including SMOTE, ADASYN, SMOTEENN, and SMOTETOMEK. Notably, the True Positive (TP) values exhibit a decrease compared to the results obtained with under-sampling techniques. Despite this reduction in TP values, it is crucial to observe that the False Positive (FP) values have also shown a decrease when contrasted with the figures achieved through under-sampling methods.

Table \ref{tab:tpfpwithoversampling} demonstrates that FP rates have decreased for both individual and ensemble ML classifiers. However, the attained TP rates have reduced. Among all classifiers, AdaBoost (AdB) attains the highest TPR score of 0.59 with the SMOTETOMEK sampling technique, while the Ensemble Hard-Voting (EHV) classifier secures the lowest FPR value of 0.03 i.e. only 3\% across all over-sampling methods. The remaining ML classifiers, including Ensemble-Stacked (ES) and Ensemble Soft-Voting (ESV), achieve FPR scores of either 0.04 or 0.05, except for the Decision Tree (DT), Gradient Boosting (GB), and AdaBoost which record the higher FPR values between 0.06 and 0.08. Moreover, All the ML classifiers have achieved an average TPR value of about 50\% except the Random Forest (RF) which scores the lowest TP rate of 0.23. 

Turning to ROC-AUC scores with oversampling techniques, Table \ref{tab335} displays the performance of all ML classifiers. Among all classifiers, AdaBoost secures the highest ROC-AUC values, ranging from 0.80 to 0.85. Conversely, the DT classifiers exhibit the lowest ROC-AUC scores with oversampled data. The ADASYN over-sampling technique demonstrates superior performance compared to other sampling methods, yielding favorable ROC-AUC values for all ML classifiers within the range of 0.71 to 0.83. overall, the ADASYN over-sampling technique shows a better performance compared to other over and combined sampling methods.

\begin{table*}[h]
\caption{Comparison between TPR and FPR values of ML classifiers after Over-Sampling}
\centering
\begin{scriptsize}
\begin{tabular*} {\linewidth}{l @{\extracolsep{\fill}} c c c c c c c c c}
\hline
\multicolumn{1}{l}{} & \multicolumn{2}{c}{SMOTE} & \multicolumn{2}{c}{ADASYN} & \multicolumn{2}{c}{SMOTEENN} & \multicolumn{2}{c}{SMOTETOMEK} \\ 
\cline{2-9}& TPR & FPR & TPR & FPR & TPR & FPR & TPR & FPR\\\hline
DT & 0.41 &  0.06 & 0.41 &  0.06 & 0.36 &  0.06 & 0.41 &  0.05\\
GB & 0.45 &  0.07 & 0.55 &  0.06 & 0.59 &  0.07 & 0.50 &  0.07\\
RF & 0.23 &  0.04 & 0.32 &  0.04 & 0.36 &  0.04 & 0.23 &  0.04\\
AdB & 0.59 &  0.08 & 0.59 &  0.08 & 0.55 &  0.08 & 0.59 &  0.08\\
ES & 0.32 &  0.04 & 0.41 &  0.05 & 0.36 &  0.04 & 0.23 &  0.04\\
EHV & 0.36 &  0.03 & 0.36 &  0.03 & 0.36 &  0.04 & 0.36 &  0.03\\
ESV & 0.27 &  0.04 & 0.36 &  0.04 & 0.36 &  0.04 & 0.32 &  0.04\\
\hline
\end{tabular*}
\end{scriptsize}
\label{tab:tpfpwithoversampling}
\end{table*}

\begin{table*}[ht]
\begin{center}

\caption{ROC-AUC Scores after Over-Sampling the data}\label{tab335}%
\begin{tabular*} {\linewidth}{l @{\extracolsep{\fill}} c c c c c}
\toprule
Classifiers & SMOTE & ADASYN & SMOTEENN & SMOTETOMEK\\
\midrule
DT & 0.59 & 0.58 & 0.56 & 0.59\\
GBoost & 0.77 & 0.79 & 0.78 & 0.84\\
RF & 0.71 & 0.71 & 0.71 & 0.71\\
AdaBoost & 0.83 & 0.83 & 0.80 & 0.85\\
Ensemble (Stacked) & 0.63 & 0.75 & 0.75 & 0.65\\
Ensemble (Hard-Voting) & 0.63 & 0.75 & 0.75 & 0.6\\
Ensemble (Soft-Voting) & 0.63 & 0.75 & 0.75 & 0.6\\
\bottomrule
\end{tabular*}
\end{center}
\end{table*}
\subsection{Comparative Analysis between Undersampling and Oversampling methods}\label{usvsos}
Both under-sampling and over-sampling techniques exert distinct effects on Blockchain anomaly detection. Under-sampling methods address the minority class instances by selecting an equivalent number of instances from the majority class through various distance-based techniques \cite{liu2008exploratory}. Conversely, over-sampling methods originate from the majority class instances and craft an equal number of synthetic instances from the minority class using different distance techniques \cite{gosain2017handling}. Both sampling techniques yield varying impacts on the classification of Bitcoin transaction data. A comprehensive comparative analysis is detailed from Table \ref{tab:tpr} to Table \ref{tab:overall}. The confusion matrices in figure \ref{cm--with-under-sampling} and \ref{cm-with-over-sampling} suggest a nuanced trade-off between true positives and false positives, emphasizing the importance of a comprehensive evaluation of the model's performance under different sampling strategies.  

Table \ref{tab:tpr} showcases that TPR scores are relatively higher for under-sampling methods such as Random Under Sampling (RUS) and the proposed XGBCLUS method, compared to over-sampling and combined sampling methods. Conversely, over-sampling and combined techniques surpass under-sampling methods in terms of FPR scores. Among under-sampling methods, XGBCLUS secures the highest TPR value of 0.91, which also stands as the pinnacle across all under-sampling, over-sampling, and combined techniques. Among over and combined balancing methods, the highest TPR value is 0.59 achieved through SMOTETOMEK. Under-sampling methods prove more effective than over-sampling methods in increasing True Positive (TP) rates. With Under-sampling methods, Machine Learning classifiers can accurately identify a larger portion of anomalous transactions.

In terms of FPR scores as shown in Table \ref{tab:fprcomp}, the best value of 0.03 is achieved across all over-sampling and combined methods. However, under-sampling methods yield a modest FPR score of 0.14 with XGBCLUS. Over-sampling methods prove more effective than under-sampling methods in reducing False Positive (FP) rates. With over-sampling methods, Machine Learning classifiers can accurately identify a larger portion of non-anomalous transactions. 

\begin{figure*}[h]
     \begin{subfigure}[b]{1\textwidth}
         \centering
         \includegraphics[width=5.5in,height=2.8in]{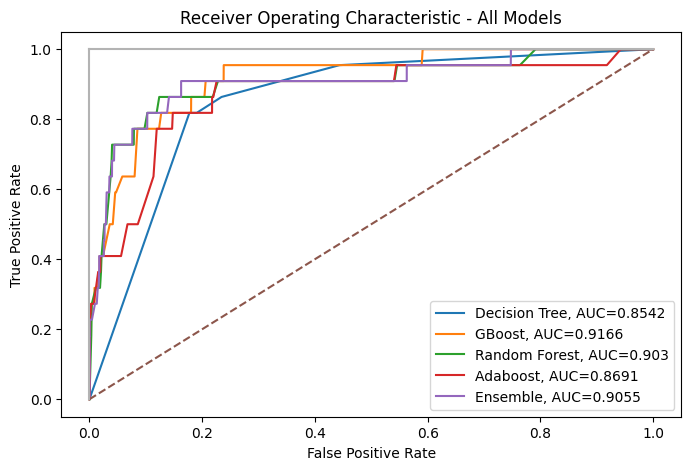}
         \caption{}
         \label{rocwithundersampling}
     \end{subfigure}\\[5mm]
     \begin{subfigure}[b]{1\textwidth}
         \centering
         \includegraphics[width=5.5in,height=2.8in]{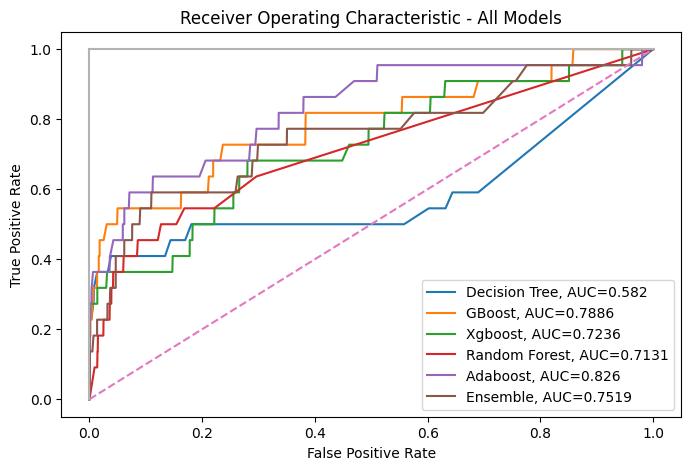}
         \caption{}
         \label{rocwithoversampling}
     \end{subfigure}
    \caption{ ROC-AUC curves for all models after (a) Under-Sampling with XGBCLUS, (b) Over-Sampling with ADASYN}
    \label{roc-curves-with-under-over-sampling}
\end{figure*}

Figure \ref{roc-curves-with-under-over-sampling} depicts that all ML classifiers exhibit improved ROC-AUC scores using the XGBCLUS under-sampling method, outperforming the scores achieved through the ADASYN over-sampling technique. This suggests that under-sampling methods excel over over-sampling methods in terms of ROC-AUC scores.
\begin{table*}[h]
\caption{TPR or Sensitivity of ML classifiers after Under-Sampling and Over-Sampling}
\centering
\begin{scriptsize}
\begin{tabular*} {\linewidth}{l @{\extracolsep{\fill}} c c c c c c c}
\hline
\multicolumn{1}{l}{} & \multicolumn{2}{c}{Under Sampling} & \multicolumn{2}{c}{Over Sampling} & \multicolumn{2}{c}{Combined Sampling}\\ 
\cline{2-7}& RUS & XGBCLUS & SMOTE & ADASYN & SMOTEENN & SMOTETOMEK \\\hline
DT & 0.64 &  0.82 & 0.41 &  0.41 & 0.36 &  0.41 \\
GB & 0.64 &  0.86 & 0.45 &  0.55 & 0.59 &  0.55\\
RF & 0.68 &  0.86 & 0.23 &  0.32 & 0.36 &  0.23\\
AdB & 0.72 &  0.81 & 0.59 &  0.59 & 0.55 &  0.64\\
ES & 0.64 &  0.86 & 0.32 &  0.41 & 0.36 &  0.23\\
EHV & 0.64 &  0.81 & 0.36 &  0.36 & 0.36 &  0.36\\
ESV & 0.68 &  0.91 & 0.27 &  0.36 & 0.36 &  0.32\\
\hline
\end{tabular*}
\end{scriptsize}
\label{tab:tpr}
\end{table*}

\begin{table*}[h]
\caption{FPR of ML classifiers after Under-Sampling and Over-Sampling}
\centering
\begin{scriptsize}
\begin{tabular*} {\linewidth}{l @{\extracolsep{\fill}} c c c c c c c}
\hline
\multicolumn{1}{l}{} & \multicolumn{2}{c}{Under Sampling} & \multicolumn{2}{c}{Over Sampling} & \multicolumn{2}{c}{Combined Sampling}\\ 
\cline{2-7}& RUS & XGBCLUS & SMOTE & ADASYN & SMOTEENN & SMOTETOMEK \\\hline
DT & 0.26 &  0.18 & 0.06 &  0.06 & 0.06 &  0.05 \\
GB & 0.26 &  0.19 & 0.07 &  0.06 & 0.07 &  0.07\\
RF & 0.20 &  0.16 & 0.04 &  0.04 & 0.04 &  0.04\\
AdB & 0.26 &  0.21 & 0.08 &  0.08 & 0.08 &  0.08\\
ES & 0.22 &  0.15 & 0.04 &  0.05 & 0.04 &  0.04\\
EHV & 0.21 &  0.14 & 0.03 &  0.03 & 0.04 &  0.03\\
ESV & 0.21 &  0.17 & 0.04 &  0.04 & 0.04 &  0.04\\
\hline
\end{tabular*}
\end{scriptsize}
\label{tab:fprcomp}
\end{table*}

\subsection{Effects of Ensemble Classifiers}\label{eofec}
\begin{table*}[h]
\caption{Accuracy of Single and Ensemble classifiers after Under-Sampling and Over-Sampling}
\centering
\begin{scriptsize}
\begin{tabular*} {\linewidth}{l @{\extracolsep{\fill}} c c c c c c c}
\hline
\multicolumn{1}{l}{} & \multicolumn{2}{c}{Under Sampling} & \multicolumn{2}{c}{Over Sampling} & \multicolumn{2}{c}{Combined Sampling}\\ 
\cline{2-7}& RUS & XGBCLUS & SMOTE & ADASYN & SMOTEENN & SMOTETOMEK \\\hline
DT & 0.74 &  0.82 & 0.94 &  0.94 & 0.94 &  0.94 \\
GB & 0.74 &  0.81 & 0.93 &  0.94 & 0.93 &  0.93\\
RF & 0.80 &  0.83 & 0.96 &  0.96 & 0.96 &  0.96\\
AdB & 0.74 &  0.79 & 0.92 &  0.92 & 0.92 &  0.92\\
ES & 0.78 &  0.85 & 0.96 &  0.95 & 0.96 &  0.96\\
EHV & 0.79 &  0.86 & 0.96 &  0.97 & 0.96 &  0.97\\
ESV & 0.79 &  0.83 & 0.96 &  0.96 & 0.96 &  0.96\\
\hline
\end{tabular*}
\end{scriptsize}
\label{tab:accuracy}
\end{table*}
While a single tree-based ML classifier can identify both anomalous and non-anomalous transactions, stacked and voting classifiers have been implemented to mitigate issues related to errors and overfitting. As demonstrated by the experimental results in Table \ref{tab:overall}, ensemble classifiers exhibit superior performance compared to individual tree-based models, particularly for accuracy and FPR values.

Table \ref{tab:accuracy} showcases the enhanced test accuracy of the three ensemble methods in comparison to single classifiers. The voting (hard) classifier attains the highest accuracy of 97\%, which stands as the peak value across all single classifiers. Among the ensemble classifiers, the voting (soft) classifier secures the highest TPR or sensitivity value of 0.91 when utilizing under-sampled data from the XGBCLUS method. Concerning the FP rate, the voting (hard) classifier achieves the best value of 0.03, capitalizing on the XGBCLUS under-sampling technique.

\begin{table*}[h]
\caption{Evaluation metrics after Under-Sampling and Over-Sampling with test data}
\centering
\begin{scriptsize}
\begin{tabular*} {\linewidth}{l @{\extracolsep{\fill}} c c c c c c c c c c c}
\hline
\multicolumn{1}{l}{} & \multicolumn{5}{c}{Under Sampling} & \multicolumn{5}{c}{Over Sampling} \\ 
\cline{2-11}& Acc & TPR/Sensitivity & FPR & ROC-AUC & Acc & TPR/Sensitivity & FPR & ROC-AUC \\\hline
DT & 0.82 &  0.82 &  0.18 & 0.85 &  0.94 &  0.41 & 0.05 & 0.58 \\
GB & 0.81 &  0.86 & 0.19 & \textbf{0.92} &  0.94 &  0.55 & 0.06 & 0.79\\
RF & 0.83 &  0.86 & 0.16 & 0.90 &  0.96 &  0.36 & 0.04 & 0.71\\
AdB & 0.79 &  0.81 & 0.21 & 0.87 &  0.92 &  \textbf{0.59} & 0.08 & \textbf{0.83}\\
ES & \textbf{0.86} &  \textbf{0.91} & \textbf{0.14} & 0.91 &  \textbf{0.97} &  0.41 & \textbf{0.03} & 0.75\\
\hline
\end{tabular*}
\end{scriptsize}
\label{tab:overall}
\end{table*}

With oversampled data, although the TPR values for the three ensemble classifiers exhibit a minor decline, the voting (hard) classifier attains the best FPR value of 0.03. Furthermore, ensemble classifiers prove effective in reducing the False Positive (FP) rate, thereby enhancing the correct identification of non-anomalous transactions. The voting (hard) classifier outperforms the other two ensemble methods by securing the highest accuracy of 0.97, alongside a commendable ROC-AUC score of 0.75.

In a comparative analysis with two prior studies of \cite{shafiq2019anomaly} and \cite{sayadi2019anomaly} that focused on blockchain anomaly detection using the Bitcoin dataset as ours, Table \ref{tabcomp} illustrates a detailed juxtaposition. Notably, our proposed ensemble method showcases superior performance across all metrics. While the studies share the primary objective of identifying malicious transactions, our emphasis extends to the accurate detection of non-malicious transactions as well. In this regard, our study endeavors to mitigate false positive rates, resulting in an enhanced true positive rate. The proposed ensemble method exhibits substantial enhancements in both true positive and false positive values, leveraging preprocessed sampled data obtained through various preprocessing steps, including feature selections. Our ensemble method excels in accuracy and notably the FPR value. A significant distinction among the studies is that our research incorporates explainability, along with the inclusion of decision rules, a component absent in their investigation.
\begin{table*}[!ht]
\begin{center}

\caption{Comparison between Proposed and Existing Work for Bitcoin anomaly detection}\label{tabcomp}%
\begin{tabular*} {\linewidth}{l @{\extracolsep{\fill}}c c c c c c c}
\toprule
References & Model & Accuracy & FPR & Explainability & Anomaly Rules\\
\midrule
\cite{shafiq2019anomaly} & Ensemble Classifiers & 0.94 & 0.05 & No & No\\
\cite{sayadi2019anomaly} & OCSVM & 0.90 & 0.09 & No & No\\
Our Study & Ensemble Classifiers & \textbf{0.97} & \textbf{0.03} & \textbf{Yes} & \textbf{Yes}\\
\bottomrule
\end{tabular*}
\end{center}
\end{table*}
\subsection{SHAP-based Explainability Analysis}\label{explain}
\begin{figure}[h]%
    \centering
    \includegraphics[width=3.4in,height=2.5in]{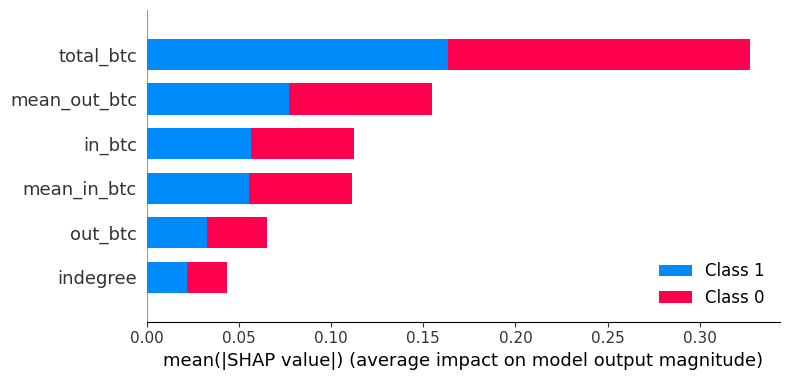}
    \caption{Hierarchy of the features contributing to classification}
    \label{fig777}
\end{figure}
SHAP is a powerful technique used to explain the predictions of machine learning models. It provides a way to attribute the contribution of each feature to a specific prediction, helping us understand why a model made a certain decision. In this study, the SHAP KernelExplainer method is employed to compute SHAP values for explaining the ensemble model's behavior, both for individual instances and the entire dataset. SHAP values are indicated by the blue color, representing their average impact on detecting anomalous transactions, while normal transactions are represented by red color SHAP values. Figure \ref{fig777} provides a hierarchical summary of features, ordered by their contributions to the model's output. The magnitude of the SHAP values indicates the strength of the feature's influence. Features with larger absolute SHAP values have a more substantial impact on the anomaly prediction. Notably, the feature "total\_btc" exerts the most substantial average impact on the model's classification, whereas "indegree" contributes the least. Furthermore, the second-highest mean SHAP value is attributed to "mean\_out\_btc," playing a significant role in anomaly detection. Conversely, "out\_btc" contributes the second least to the classification. "in\_btc" and "mean\_in\_btc" exhibit nearly equivalent contributions to the classification of Bitcoin transactions.

Figure 10 illustrates a comparison of SHAP values for four randomly selected anomalous and normal transactions, as predicted by the ensemble model post both under-sampling and over-sampling techniques. As depicted in Figure 10, the baseline value for the expected anomalous transaction is 0.49 under under-sampling and 0.50 under over-sampling. Notably, the scores for anomalous Bitcoin transactions are 0.06 and 0.40 with under-sampled and over-sampled data, respectively. These scores are notably lower than the baseline.

\begin{figure*}[H]
     \begin{subfigure}[b]{1\textwidth}
         \centering
         \includegraphics[width=4.8in,height=1.8in]{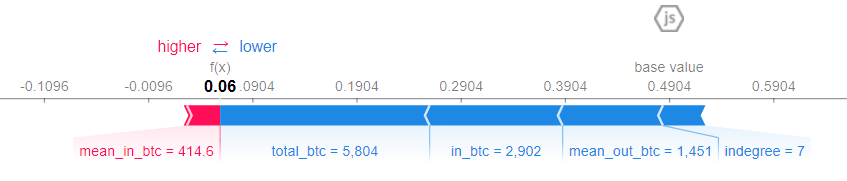}
         \caption{}
         \label{positiveundersampling}
     \end{subfigure}\\
     \begin{subfigure}[b]{1\textwidth}
         \centering
         \includegraphics[width=4.8in,height=1.8in]{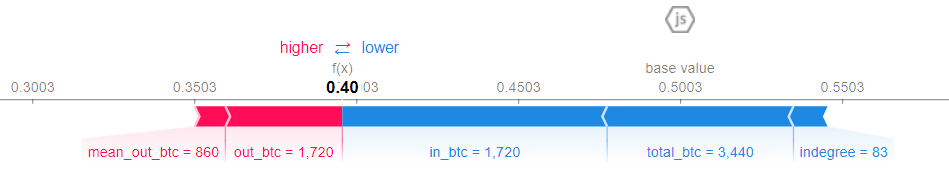}
         \caption{}
         \label{positiveoversampling}
     \end{subfigure}
     \begin{subfigure}[b]{1\textwidth}
         \centering
         \includegraphics[width=4.8in,height=1.8in]{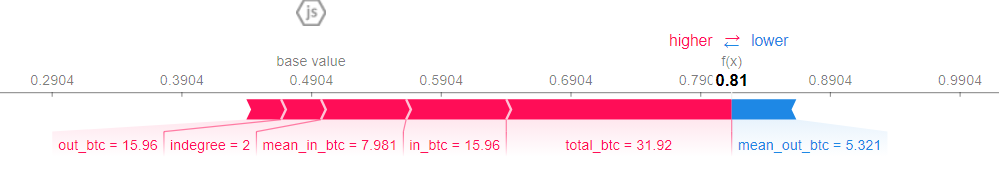}
         \caption{}
         \label{negativeundersampling}
     \end{subfigure}\\
     \begin{subfigure}[b]{1\textwidth}
         \centering
         \includegraphics[width=4.8in,height=1.8in]{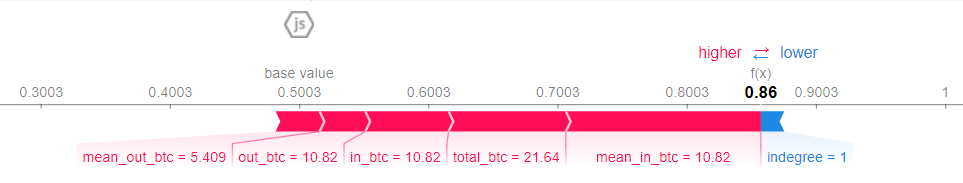}
         \caption{}
         \label{negativeoversampling}
     \end{subfigure}
    \caption{ Features contributing to (a) detect a positive case after Under-Sampling, (b) detect a positive case after Over-Sampling, (c) detect a negative case after Under-Sampling, (d) detect a negative case after Over-Sampling}
    \label{force-curve-with-under-over-sampling}
\end{figure*}

The SHAP value representations in Figures \ref{positiveundersampling} and \ref{positiveoversampling} elucidate the features contributing the most to lowering the score. These figures underscore that "total\_btc" holds the highest importance, followed by "in\_btc," "mean\_out\_btc," and "indegree." For normal Bitcoin transaction data, the anomalous transaction scores are 0.81 and 0.86 with under-sampled and over-sampled data, respectively. These scores exceed the baseline. SHAP value representations for normal transactions are displayed in Figures \ref{negativeundersampling} and \ref{negativeoversampling}. As evident from the figures, the SHAP values are generally small, except for "mean\_out\_btc" and "indegree" under under-sampling and over-sampling, respectively.

Comparing all four cases in Figure 10, it's evident that the feature values for anomalous instances are higher than those for normal instances, making them discernible to human observation. To facilitate a better human understanding, a comparison between actual anomalous and normal data instances is presented in Table \ref{table11}. This table indicates that the features identified by SHAP values in Figure 10 exhibit significant disparities in actual transaction data. Hence, it can be concluded that SHAP values are effective in explaining both normal and anomalous Bitcoin transactions.
\begin{table}[!ht]
\begin{center}
\caption{Comparison of actual values of an anomalous and normal Bitcoin transaction}\label{table11}%
\begin{tabular}{@{}llll@{}}
\toprule
Feature Name & Anomalous & Normal\\
\midrule
Indegree    & 7 & 2\\
in\_btc    & 2902 & 15.96\\
out\_btc    & 2902 & 15.96\\
total\_btc    & 5804 & 31.92 \\
mean\_in\_btc    & 414.6 & 7.98\\
mean\_out\_btc    & 1451 & 5.32\\
\bottomrule
\end{tabular}
\end{center}
\end{table}

\begin{table}[!ht]
\begin{center}
\caption{Feature Importance values}\label{tablefet}%
\begin{tabular}{@{}llll@{}}
\toprule
Feature Name & Importances\\
\midrule
total\_btc    & 0.666268\\
mean\_in\_btc    & 0.124338\\
mean\_out\_btc    & 0.092236\\
in\_btc    & 0.046185\\
out\_btc & 0.036751\\
Indegree    & 0.034222\\
\bottomrule
\end{tabular}
\end{center}
\end{table}

\subsection{Anomaly Rule Generation and Interpretability Analysis}\label{tree-explain}
\begin{figure*}[h]%
    \centering
    \includegraphics[width=0.9\textwidth]{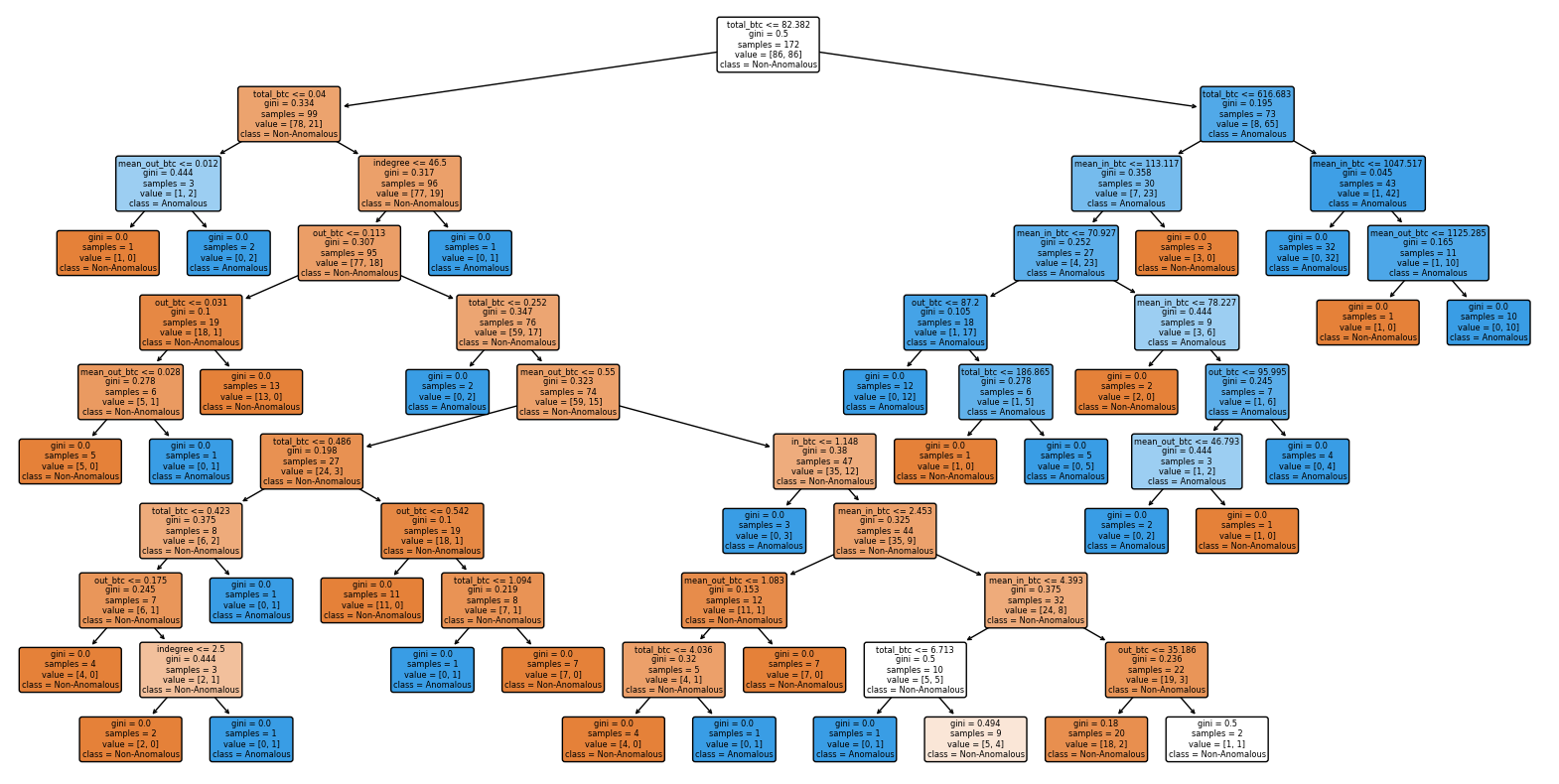}
    \caption{Visualization of DT of depth 10 for generating anomalous rules}
    \label{figDT}
\end{figure*}

Interpreting anomaly rules using tree representations can help in understanding why certain instances are classified as anomalies. It provides a step-by-step breakdown of the decision process and highlights which features and thresholds played a crucial role in making the decision. In this study, having explored tree-based machine learning classifiers for detecting anomalous Bitcoin transactions, tree visualization offers a structured and interpretable approach to comprehending how the model arrives at decisions using input features. A visualization of the decision tree with max-depth 10 is shown in Figure \ref{figDT}. The top node of the tree is called the root node, and it represents the entire dataset. Each subsequent node represents a decision point based on a specific feature and threshold. Moving down the sub-trees, each node represents a feature and a corresponding threshold. Instances are directed to different branches based on whether their feature values satisfy the given threshold. Each leaf node corresponds to a specific prediction class, in this case, anomalous or non-anomalous. 
\begin{table*}[ht]
\begin{center}
\caption{Significant Rules for being anomalous Bitcoin transaction}\label{table-rules}%
\begin{tabular*}{\linewidth}{l @{\extracolsep{\fill}} c c c c c}
\toprule
Anomaly rule & Class & Total Samples & Correctly Identified & Confidence(\%)\\
\midrule
$1.$	If (total\_btc is greater than 616.683 and\\ mean\_in\_btc is less than or equal to 1047.517) then    & Anomalous & 32 & 32 & 100\\
$2.$	If (total\_btc is greater than 616.683 and\\ mean\_in\_btc is greater than 1047.517,\\ and mean\_out\_btc is greater than 1125.285) then    & Anomalous & 10 & 10 & 100\\
$3.$	If (total\_btc is greater than 82.38 and\\ total\_btc is less or equal to 616.683 \\and mean\_in\_btc is less or equal to 70.927, \\and out\_btc is less or equal to 87.2) then    & Anomalous & 12 & 12 & 100\\
$4.$	If (total\_btc is greater than 82.38 and\\ total\_btc is less or equal to 616.683 \\and mean\_in\_btc is less or equal to 70.927) then    & Anomalous & 18 & 17 & 94\\
$5.$	If (total\_btc is greater than 616.683 and\\ mean\_in\_btc is greater than 1047.517) then    & Anomalous & 11 & 10 & 91\\
\bottomrule
\end{tabular*}
\end{center}
\end{table*}

In Table \ref{tablefet}, the descending order of feature importance is presented, with numerical values representing the total normalized reduction in Gini Impurity achieved by splitting the respective feature throughout the tree. Unsurprisingly, 'total\_btc,' the feature utilized to split at the root node, holds the highest importance. This implies that 'total\_btc,' followed by 'mean\_in\_btc',' are the most crucial features in determining the anomaly status of a Bitcoin transaction.

By traversing the decision tree from the root node to a specific leaf node, we can extract the sequence of decisions (feature and threshold combinations) that lead to an anomaly prediction. These decisions essentially form the "anomaly rules" for that instance. In Table \ref{table-rules}, we have a set of rules along with the confidence scores that lead to an anomalous node regarding the tree in Figure \ref{figDT}. Decision trees also provide a measure of feature importance. Features closer to the root of the tree have a more significant influence on the overall decision-making process. So total\_btc is the most important feature that contributes the highest to deciding whether a transaction is anomalous or not. Moreover, both mean\_in\_btc and mean\_out\_btc have valuable effects on classifying anomalous transactions. However, in some cases, the indegree may also contribute to detecting anomalous transactions.
The confidence score reflects the model's confidence or certainty in its prediction that the transaction is anomalous based on the specified conditions in the rules. Rules 1 to 3 give the highest confidence scores i.e. the highest probability for being anomalous.  However, rules 4 to 5 have an average probability for the Bitcoin transactions to be anomalous. From Table \ref{table-rules}, the more thorough the examination of decision rules, the greater the confidence in accurately identifying anomalous transactions. Although we've specifically crafted anomaly rules for our Bitcoin transaction data, this methodology holds promise for creating effective decision rules in diverse domains, particularly within the realm of Blockchain.

\section{Discussion}\label{disc}
Detecting anomalies from highly imbalanced data presents a formidable challenge. Although tree-based machine learning (ML) classifiers demonstrate effectiveness in anomaly detection, the significant data imbalance tends to bias these classifiers towards non-anomalous transactions. As demonstrated in Table \ref{tab222}, this imbalance often results in low True Positive rates for single tree-based ML classifiers. Given that accurate detection of anomalous transactions is the primary objective of this study, addressing data imbalance before model training becomes crucial.

Two main approaches for balancing data, oversampling and under-sampling techniques, are available. The selection of the appropriate technique is a critical decision for researchers, particularly when dealing with highly imbalanced data. Oversampling methods generate synthetic data to balance both majority and minority classes, but they may not effectively detect most anomalous transactions. On the other hand, under-sampling techniques aim to balance classes by focusing on minority samples, leading to high TPR and low FPR values.

For anomaly classification problems, an under-sampling technique could be suitable if it can maximize the True Positive value while minimizing the False Positive value. Regrettably, the prevalent down-sampling algorithms in existence often fall short of achieving this balance, as illustrated in the results analysis section. While the Random Under Sampling (RUS) technique can assist in achieving better class balance, it does come with drawbacks. This technique decreases the volume of training data available, potentially resulting in the loss of crucial samples and subsequently leading to suboptimal model performance. The Near-Miss-1 technique, although contributing to enhanced model performance with a TPR score of 0.95, exhibits a significant False Positive rate. Recognizing the limitations of RUS and Near-Miss-1, we have introduced the XGBCLUS under-sampling technique to address these issues. XGBCLUS surpasses its predecessors in detecting anomalous Bitcoin transactions. Both individual and ensemble classifiers demonstrate superior TPR, FPR, and ROC-AUC values when trained on under-sampled data using XGBCLUS, as demonstrated in Section \ref{eofus}.

Furthermore, our investigation extends to over-sampling and combined data balancing techniques, aiming to highlight the distinctions between under-sampling and over-sampling approaches. Comparative results in Section \ref{usvsos} reveal that under-sampling methods are effective in improving the True Positive (TP) value but lead to a higher False Positive (FP) rate. Conversely, over-sampling methods contribute to achieving an increase in TP and a decrease in FP compared to unbalanced data. Among the various over and combined sampling techniques, ADASYN outperforms SMOTE, while SMOTEENN proves superior to SMOTETOMEK.       

In the realm of classification, we introduce both stacked and voting (hard and soft) ensemble classifiers in addition to individual tree-based ML classifiers. It's evident that ensemble classifiers excel over individual ML classifiers when applied to both under and over-sampled data. Among the three ensemble classifiers, the Hard-Voting classifier demonstrates superior performance, achieving the highest TPR and lowest FPR values for both under and over-sampled data. In situations involving imbalanced data, the FPR assumes significance as an essential evaluation metric. The ensemble classifiers consistently yield improved FPR values compared to single ML classifiers, emphasizing the effectiveness of ensemble approaches in handling imbalanced data.

Nonetheless, machine learning models often act as opaque "Black Boxes". In order to substantiate human assumptions and model predictions, we delve into the realm of Explainable AI using SHAP, a facet elaborated in section \ref{explain}. Aggregating the SHAP values across the entirety of dataset instances enables the identification of the most pivotal feature, which, in this context, is "total\_btc". This salient feature's importance is also discernible to human observers. Furthermore, we expound upon anomaly rules derived from Decision Trees (DT). As illustrated in Figure \ref{tree-explain}, "total\_btc" emerges as the root node, signifying its paramount role in classifying anomalous transactions. Both SHAP's analysis and the Tree representation collectively pinpoint "total\_btc" as the quintessential feature contributing to the identification of Bitcoin anomalous transactions. Overall, SHAP provides a transparent and interpretable way to understand how each feature contributes to the anomaly detection process. This can help data analysts and domain experts identify patterns, correlations, and potential anomalies in the data that the model is leveraging for its predictions. In addition, Interpreting anomaly rules using tree representations can help in understanding why certain instances are classified as anomalies. It provides a step-by-step breakdown of the decision process and highlights which features and thresholds played a crucial role in making the decision. This transparency is especially valuable in domains where explainability is important, allowing stakeholders to validate the model's decisions and identify potential issues in the data or model.

The findings of our study bear significant implications for enhancing both blockchain security and anomaly detection. The efficacy of the proposed ensemble model in detecting anomalous transactions in Bitcoin signifies a promising approach to enhance the security of blockchain systems. Additionally, the incorporation of eXplainable Artificial Intelligence (XAI) techniques, including SHAP analysis, along with anomaly rules for interpretability analysis, contributes transparency and clarity to the anomaly detection process. Implementing such explainability measures can contribute to robust anomaly detection systems and, consequently, fortify the overall security of blockchain technologies. Additionally, the exploration of various sampling techniques, including the proposed under-sampling and combined-sampling methods, contributes to the development of strategies for handling highly imbalanced datasets in the blockchain domain. This has broader implications for anomaly detection in other contexts where imbalanced data is a common challenge. 

In future research endeavors, there is potential to delve into the efficacy of our proposed under-sampling techniques, alongside over-sampling methods, for anomaly detection in different Blockchain domains such as Ethereum transactions \cite{tikhomirov2018ethereum}, credit card fraud detection \cite{dornadula2019credit}, and money laundering \cite{chen2018machine}. The exploration of a data-driven approach \cite{sarker2021data} could lead to the development of real-time blockchain anomaly detection systems. In future research, we plan to gather additional data related to blockchain and perform further experimental analyses. Furthermore, a comparative analysis between Machine Learning and Deep Learning algorithms could be undertaken to determine the optimal model for anomaly detection in Blockchain transactions. This pursuit holds promise for enhancing the precision and effectiveness of anomaly detection methodologies.
\section{Conclusion}\label{cons}
In this study, we have conducted an extensive comparative analysis aimed at detecting anomalies within Blockchain transaction data. While numerous studies have been conducted in this field, a prevailing limitation has been the absence of explanations for model predictions. To address this shortcoming, our study endeavors to combine eXplainable Artificial Intelligence (XAI) techniques and anomaly rules with tree-based ensemble classifiers using Bitcoin transactions. Notably, the Shapley Additive exPlanation (SHAP) method plays a pivotal role in quantifying the contribution of each feature toward predicting the model's output. Moreover, the anomaly rules help to interpret whether a Bitcoin transaction is anomalous or not. Consequently, one can readily identify which features play a pivotal role in anomaly detection. To further enhance our methodology, we have developed an under-sampling algorithm termed XGBCLUS. This algorithm facilitates the balance between anomalous and non-anomalous transaction data, and its performance has been compared against other well-known under-sampling and oversampling techniques. Subsequently, the results derived from various single tree-based classifiers are juxtaposed against those obtained from stacking and voting ensemble classifiers. Our findings unequivocally demonstrate that our proposed under-sampling method, XGBCLUS, has yielded improvements in TPR and ROC-AUC scores. Additionally, ensemble classifiers have exhibited superior performance in comparison to popular single ML classifiers. In conclusion, our study highlights that ensemble classifiers, when combined with suitable balancing techniques, prove effective in detecting Blockchain anomalous transactions.
\printcredits

\bibliographystyle{cas-model2-names}

\bibliography{ref}

\end{document}